\documentclass[10pt, a4paper]{article}

\usepackage[T1]{fontenc}

\usepackage{svg}
\usepackage{float}
\usepackage{listings, lstautogobble}
\usepackage{multirow}
\usepackage{booktabs}
\usepackage[hidelinks]{hyperref}
\newcommand*\rot{\rotatebox{90}}
\usepackage{lrec-coling2024} % this is the new style

\title{{NLPre}: a revised approach towards language-centric benchmarking\\of {N}atural {L}anguage {Pre}processing systems}%.\\The Polish scenario}

\name{Martyna Wiącek, Piotr Rybak, Łukasz Pszenny, Alina Wróblewska} 

\address{Institute of Computer Science, Polish Academy of Sciences \\
         ul. Jana Kazimierza 5, 01-248 Warsaw, Poland \\
         \{m.wiacek, p.rybak, l.pszenny, alina\}@ipipan.waw.pl}
\abstract{
With the advancements of transformer-based architectures, we observe the rise of natural language preprocessing (\hbox{NLPre}) tools capable of solving preliminary NLP tasks (e.g. tokenisation, part-of-speech tagging, dependency parsing, or morphological analysis) without any external linguistic guidance. It is arduous to compare novel solutions to well-entrenched preprocessing toolkits, relying on rule-based morphological a\-na\-ly\-sers or dictionaries. 
Aware of the shortcomings of existing \hbox{NLPre} evaluation approaches, we investigate a novel method of reliable and fair evaluation and performance reporting. 
Inspired by the GLUE benchmark, the proposed language-centric benchmarking system enables comprehensive ongoing evaluation of multiple NLPre tools, while credibly tracking their performance. 
The prototype application is configured for Polish and integrated with the thoroughly assembled NLPre-PL benchmark. % (\href{https://nlpre-pl.clarin-pl.eu}{https://nlpre-pl.clarin-pl.eu}). 
Based on this benchmark, 
we conduct an extensive evaluation of a variety of Polish NLPre systems. 
To facilitate the construction of benchmarking environments for other languages, e.g. NLPre-GA for Irish or NLPre-ZH for Chinese, we ensure full customization of the publicly released source code of the benchmarking system. The links to all the resources (deployed platforms, source code, trained models, datasets etc.) can be found on the project website: \href{https://sites.google.com/view/nlpre-benchmark}{https://sites.google.com/view/nlpre-benchmark}.
 \\ \newline  }
\begin{document}

\maketitleabstract

\begin{figure*}[ht!]
\centering
\includegraphics[scale=0.41]{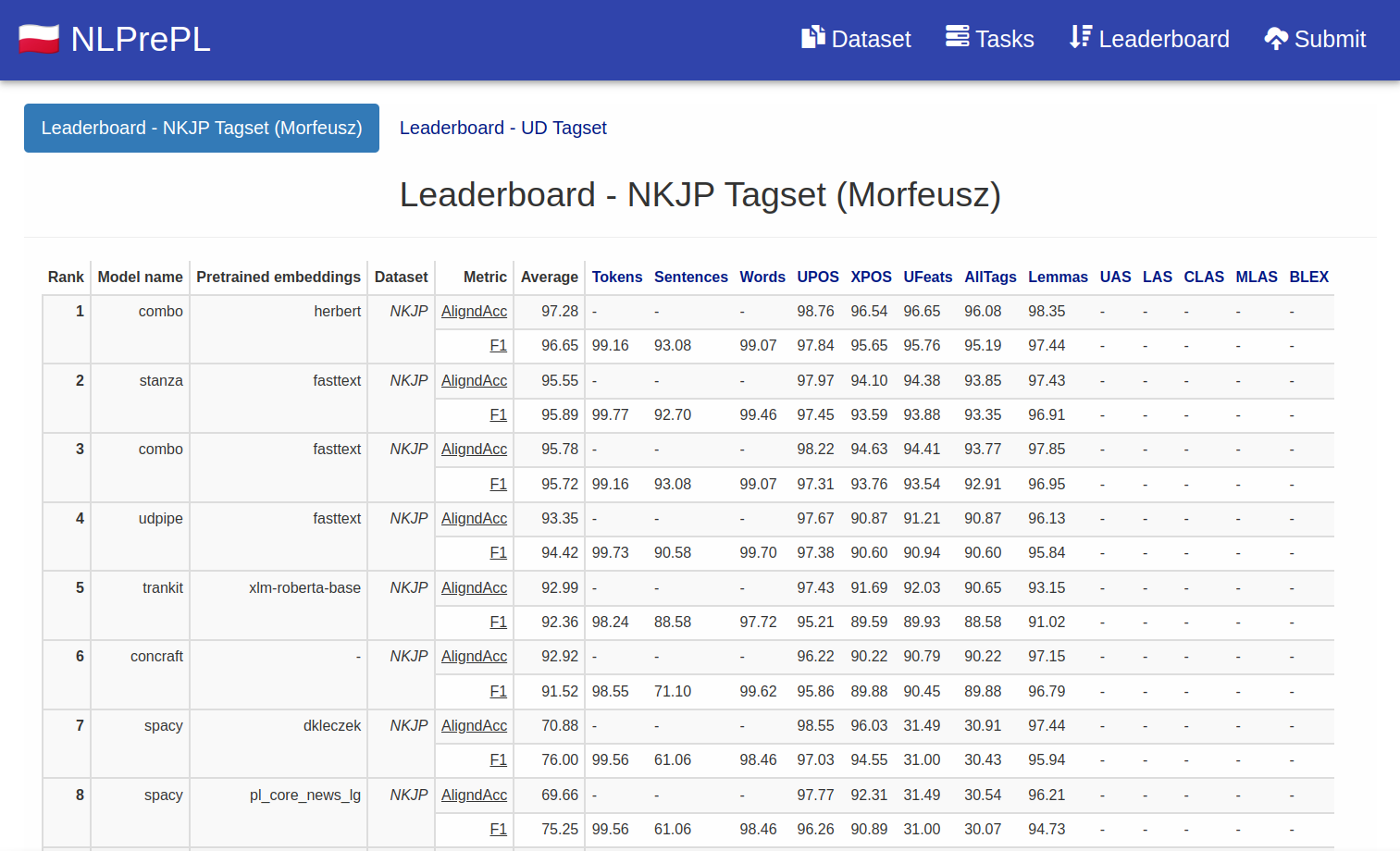}
\vspace*{-3mm}
%\caption{\small Screenshot of Polish version for NLPre benchmark.}
\caption{\small Screenshot of the NLPre-PL leaderboard.}
\label{polish_nlpre}
\vspace*{-3mm}
\end{figure*}

\section{Introduction and related works}
\label{sec:intro}
%The traditional page limit for ECAI long papers is {\bf 7 (six)} pages
%in the required format. The traditional page limit for short
%submissions is {\bf 2} pages.
%
%However, these page limits may change from one ECAI to
%another. Consult the most recent Call For Papers (CFP) for the most
%up-to-date page limits.

Morphosyntactic features predicted by part-of-speech (POS) taggers and dependency parsers underlie various downstream tasks, including but not limited to sentiment analysis \cite{sun-etal-2019-aspect}, relation extraction \cite{zhang-etal-2018-graph,vashishth-etal-2018-reside,guo-etal-2019-attention}, semantic role labelling \cite{wang-etal-2019-best,kasai-etal-2019-syntax}, question answering \cite{Khashabi_Khot_Sabharwal_Roth_2018}, or machine translation \cite{Chen_Zhao_Yang_Liu_2017,zhang-etal-2019-syntax-enhanced}. These underlying tasks may therefore be referred to as \textit{natural language preprocessing} (NLPre) tasks, as they precede the advanced NLP tasks. Since the quality of morphosyntactic predictions has a~crucial impact on the performance of downstream tasks \cite{sachan-etal-2021-syntax}, it is prudent to employ the best existing NLPre tools to predict the proper linguistic features. We are equipped with various \mbox{NLPre} methods, ranging from rule-based tools with hand-crafted grammars \cite[e.g.][]{xle}, through statistical systems \cite[e.g.][]{nivre-2009-non,mcdonald-etal-2005-online,straka-etal-2016-udpipe}, neural systems supported by pre-trained language models \cite[e.g.][]{qi-etal-2020-stanza,nguyen-etal-2021-trankit} %,klimaszewski-wroblewska-2021-combo-state}.
to large language models \cite[LLM][]{ouyang2022training}. 

In the context of intrinsically evaluating NLPre tools and reporting their performance, a~variety of approaches have been proposed, e.g. shared task, performance table, and progress repository. The main goal of a~\textit{shared task} is to comprehensively evaluate participating systems on the released datasets using the carefully defined evaluation methodology. Numerous NLPre shared tasks have been organised so far \cite[e.g.][]{buchholz-marsi-2006-conll,seddah-etal-2013-overview,%seddah-etal-2014-introducing,
zeman-etal-2017-conll,zeman-etal-2018-conll}, and they undoubtedly boosted the development of NLPre. 
While widely favoured, shared tasks are questionable as a complete and up-to-date source of knowledge about NLPre progress. First, they scrutinise only solutions propounded in the current contest and do not include systems participating in the previous editions or possible future ones. Second, as shared tasks are organised sporadically, their results are not revised and may quickly become outdated. Certainly, the datasets released for shared tasks can be reused in experiments involving novel tools. The results of such experiments can be reported in independent scientific publications. Nonetheless, %these experiments frequently restrict their scope to a subset of widely spoken languages, overlooking certain languages featured in shared tasks. Moreover, 
these publications are widely scattered, lacking a centralised platform for systematically tracking the ongoing NLPre progress with respect to a particular language. %Furthermore, this conventional approach tends to optimise models for shared task data solely to surpass shared task models, rather than pursuing a broad model optimisation approach designed to function effectively with unfamiliar data.

% Performance tables -- unpublished results, many languages but only one system (no comparison with other systems)

The results of a new %developed 
or upgraded NLPre tool are typically reported in \textit{performance tables}
 % Modern NLPre tools are constantly developed, their models are estimated on updated datasets, e.g. Universal Dependencies (UD) releases, and their results are commonly reported in performance tables 
(e.g. Stanza\footnote{\href{https://stanfordnlp.github.io/stanza/performance.html\#system-performance-on-ud-treebanks}{https://stanfordnlp.github.io/stanza/performance.html} (UD v2.8)} 
or Trankit\footnote{\href{https://trankit.readthedocs.io/en/latest/performance.html\#universal-dependencies-v2-5}{https://trankit.readthedocs.io/en/latest/performance.
html\#universal-dependencies-v2-5} (UD~v2.5)}). 
Such tables provide information about the quality of %a given 
the tool in preprocessing a set of languages. The performance tables, however, often lack comparison with other systems trained for these particular languages.
Additionally, as NLPre systems may be trained on different dataset releases (e.g. of Universal Dependencies), comparing their performance tables is not conclusive.

% Papers with code -- published results, outdated results, selected languages
Information about trends and progress in NLP research is usually collected in public repositories such as \textit{Papers with Code}\footnote{\href{https://paperswithcode.com}{https://paperswithcode.com}} or \textit{NLP-progress}\footnote{\href{http://nlpprogress.com}{http://nlpprogress.com}}. These repositories contain a~repertoire of datasets for common NLP tasks, e.g. dependency parsing and POS tagging, and rankings of models trained and tested on these datasets. They are open to contributing new datasets and results, which, to ensure their credibility, originate from published and linked scientific papers. However, cutting-edge yet unpublished results of a~new or upgraded NLPre system are not eligible to report. NLPre tasks are accompanied by datasets mostly in English, raising the problem of language unrepresentation of the repositories. Last but not least, the Papers with Code repository is prone to abuse. After logging in, one can add new results and link them with irrelevant papers as well as edit existing results. The fraudulent results are publicised immediately.

%One can consider dynamic dataset creation platforms such as Dynabench \cite{kiela-etal-2021-dynabench}, which enables users to expand the benchmark by inputting custom test examples. However, in the case of NLPre, this approach seems futile as expert morphosyntactic knowledge is essential to avoid gathering noisy examples, and finding multiple experts among casual users can be an obstacle.

Despite yielding valuable information about the progress in NLPre, the mentioned evaluation approaches also reveal shortcomings, e.g. outdated and incomplete outcomes, lack of cross-system comparison, disregarding some systems, risk of result manipulation and absence of a language-centring perspective.
%TODO - add language-centric motivation

Following standard procedures in NLP research, we propose to robustly and fairly evaluate %existing and forthcoming 
\mbox{NLPre} tools using the benchmarking method that allows for the evaluation of NLP models' performance and progress. NLP benchmarks are coupled with leaderboards that report and update model performance on the benchmark tasks, e.g. GLUE \cite{wang-etal-2018-glue}, XTREME \cite{pmlr-v119-hu20b}, GEM \cite{gehrmann-etal-2021-gem}. The conventional benchmarking approach may be dynamically enhanced, exemplified by the \textit{Dynabench} platform \cite{kiela-etal-2021-dynabench}, which enables users to augment the benchmark data by inputting custom %test 
examples. This human-and-model-in-the-loop
benchmarking scenario appears promising for NLU tasks. Nevertheless, it 
may not be effective in the case of NLPre, as annotating credible examples of syntactic trees or morphological features requires expert %morphosyntactic 
knowledge. Finding multiple experts among casual users can be a serious obstacle, we thus implement our system in tune with the %established 
standard benchmarking method.

  To our knowledge, benchmarking hasn't been used to rank NLPre systems, even if it is valuable and desired by the community creating treebanks or designing advanced NLP pipelines. Our NLPre benchmarking approach fills this~gap.
The proposed online benchmarking system automatically assesses submitted predictions of %the~evaluated 
NLPre systems and publishes their performance ranking on a~public scoreboard (see Section~\ref{sec:system}). The system is language-centric and tagset-agnostic, enables comprehensive and credible evaluation %(i.e. the test set is classified, each model is eligible to test, and results cannot be manipulated) 
 and constitutes an~up-to-date source of information on NLPre progress for a~particular language. 
 Unlike similar platforms, e.g. Codalab \cite{codalab_competitions}, the NLPre benchmarking system is fully configurable and easy to set up, allowing users to establish an evaluation environment for any language. Additionally, it can be self-hosted, making it convenient for developers and researchers working with a particular language to have it accessible on a local server.

To justify the use of the benchmarking technique for NLPre tasks, we conduct empirical research in a~challenging %language 
scenario with Polish as an~example language. In the case of Polish, one dominant hurdle arises -- the discrepancies between different tagsets, %syntactic 
annotation schemes and datasets utilised for training disparate systems preclude their direct comparison. 
We thus standardise the training and evaluation of \mbox{NLPre} systems on a new performance benchmark for Polish, hereafter NLPre-PL (see Section \ref{sec:nlpre_pl}). It consists of a~predefined set of NLPre tasks and reformulated versions of existing Polish datasets. % with training, development, and test data sampled from the same distribution. 
Section~\ref{sec:evaluation} outlines our robust and reliable evaluation of the selected NLPre systems on the NLPre-PL benchmark. According to our knowledge, no evaluation experiments have been carried out in Polish to compare the performance of off-the-shelf LLMs, 
neural NLPre systems and established tagging disambiguators due to the lack of a~coherent evaluation environment.

This work makes a tripartite contribution encompassing novelty, research, and development underpinned by an open-source ethos. (1)~We propose a novel language-oriented benchmarking approach to evaluate and rank NLPre systems. (2) We conduct a scientific evaluation of the proposed approach in the non-trivial Polish language scenario on the assembled NLPre-PL benchmark. (3) We publish online benchmarking platforms for three distinct languages: Polish\footnote{\href{https://nlpre-pl.clarin-pl.eu}{https://nlpre-pl.clarin-pl.eu}}, Chinese\footnote{\href{https://nlpre-zh.clarin-pl.eu}{https://nlpre-zh.clarin-pl.eu}}, and Irish\footnote{\href{https://nlpre-ga.clarin-pl.eu}{https://nlpre-ga.clarin-pl.eu}}, and release the benchmarking system's source code as open-source.%\footnote{\href{https://gitlab.clarin-pl.eu/taggers-evaluation/nlpre-benchmark-code}{https://gitlab.clarin-pl.eu/taggers-evaluation/nlpre-benchmark-code}}

 \iffalse
 (2) We provide an~online benchmarking platform.\footnote{%For anonymity reasons, the 
The NLPre-PL URL will be added in the final version.} % for Polish. 
 (3) We publish the code 
 % of the benchmarking system
 as open source.\footnote{%A~.zip package with the source code is available under 
 An anonymised version of the benchmarking system source code that will be distributed under GNU GPLv3 license: \href{https://anonymous.4open.science/r/code-781E/}{https://anonymous.4open.science/r/code-781E/}. %The code for the benchmarking system will be distributed under GNU GPLv3 license and the URL will be added in the final version.
 } (4) We assemble the~new NLPre-PL benchmark\footnote{A currently anonymised version of the NLPre-PL dataset: 
\href{https://anonymous.4open.science/r/dataset-3AEB/}{https://anonymous.4open.science/r/dataset-3AEB/}} and use it to benchmark %existing 
 contemporary NLPre systems. 
 \fi

\section{NLPre benchmarking}
\label{sec:nlpre}

\subsection{Research concept}
In this study, we introduce a novel adaptation of the benchmarking approach to NLPre. The primary objective is to establish an automated and credible method for evaluating NLPre systems against a provided benchmark and continuously updating their performance ranking on a~publicly accessible scoreboard. More specifically, predictions for the benchmark test sets output by NLPre systems and submitted to the benchmarking system are automatically compared against the publicly undisclosed reference %gold standard 
dataset. %, which is publicly undisclosed. 
This method effectively prevents result manipulation and ensures fairness of the final assessment. 
The second important methodological assumption is to enable the ongoing evaluation of new or upgraded NLPre systems to guarantee up-to-date and complete ranking. Consequently, the leaderboard can serve as a reliable point of reference for NLPre system developers. % (also allowing them to evaluate their systems without having to publish the results).

Based on these assumptions, we design and implement the language-centric and tagset-agnostic benchmarking system that enables comprehensive and credible evaluation, constitutes an~up-to-date source of information on NLPre progress, and is fully configurable to facilitate building benchmarking systems for multiple languages.

\subsection{Online benchmarking system}
\label{sec:system}
The benchmarking system comprises three main 
parts: a~data repository, a~submission and evaluation system, and a~leaderboard. 
The data repository provides descriptions of NLPre tasks, datasets, and evaluation metrics, as well as links to the datasets. 

The model submission and evaluation system allows the researchers to evaluate a new model by
submitting its predictions for the test sets of raw sentences. It is mandatory to upload predictions for all provided test sets for a given tagset; however, it is possible to participate in an evaluation for only one tagset and only for a selected range of tasks. %It must be considered that missing results will be automatically assigned a zero value.

The leaderboard is a tabular display of the performance of all submissions with their results for each dataset and tagset. The results for the evaluated model and its rank are displayed in the leaderboard provided the submitter confirms their publication.

The benchmarking system is implemented as a web-based application in Python using Django framework. This framework allows quite an easy implementation of MVC design pattern. Moreover, it offers access to the administrator panel, which can be very useful in the custom configuration of the benchmark. The submission scores are stored in a local SQLite database and the submissions are stored in \texttt{.zip} files in a designated directory. The results from the leaderboard are conveniently accessible via an API.

\subsection{Configuration}
\label{sec:config}
We acknowledge the need to configure similar evaluation environments for other languages to promote linguistic diversity within the worldwide NLP community and to support local NLP communities working on a~particular language. %We provide the code in an~open-sourced version in the hope that researchers of different languages will contribute to this effort and maintain the online benchmarking systems for their language.
%To demonstrate the easy of configuring our open-sourced system, we provide detailed instructions in Appendix~\ref{apx:leaderboard_config}. 
To ensure that, we publish a \texttt{.yaml} file that enables easy management of datasets, tagset, and metrics included in the benchmark. The content of all subpages can be modified using a WYSIWYG editor within the application. 
 This setting ensures quite a low entry level for setting up the platform, with minimal changes required.

%Since we propose an evaluation platform for NLPre tasks, 
As a standard feature, we include pre-defined descriptions for the prevalent NLPre tasks. Those can be modified via either configuration files or the administrator panel. Additionally, we supply a default evaluation script, but users are free to provide their own customised code.

To show the capabilities of the benchmarking system, %in this work, 
we set up a prototype for Polish (Figure~\ref{polish_nlpre}). NLPre-PL is described in detail in Section \ref{sec:nlpre_pl}. To support our claim that the system is language agnostic, we set up NLPre-GA for Irish and NLPre-ZH for Chinese. % (see Appendix~\ref{apx:nlpre_irish_chinese}). 
The choice of those languages is not arbitrary; our objective is to demonstrate the capability of the platform in evaluating diverse languages, including those based on non-Latin scripts. In setting up said benchmarking systems we use existing UDv2.9 treebanks: UD\_Chinese-GSD \citelanguageresource{ChineseTreebank} %\footnote{\href{https://github.com/UniversalDependencies/UD\_Chinese-GSD}{https://github.com/UniversalDependencies/UD\_Chinese-GSD}} 
and UD\_Irish-IDT \citelanguageresource{IrishTreebank} %\footnote{\href{https://github.com/UniversalDependencies/UD_Irish-IDT}{https://github.com/UniversalDependencies/UD\_Irish-IDT}}) 
and available up-to-date models, trained on these treebanks. The selection of models mirrors the criteria applied in this work regarding the evaluation of Polish, that is: COMBO, Stanza, SpaCy, UDPipe, and Trankit. If the specific model is not available for UDv2.9, we train it from scratch on the datasets linked above. 
%We upload the results to a suitably configured platform, obtaining leaderboards for Chinese (Figure \ref{chinese_nlpre}) and Irish (Figure \ref{irish_nlpre}). %The configuration time for each platform is in line with the run times presented in Appendix \ref{apx:leaderboard_config}.

%\section{Evaluation system}
\section{NLPre-PL benchmark}
\label{sec:nlpre_pl}

%We select Polish to verify our approach and construct a prototype of our benchmarking system. %This section introduces the \hbox{NLPre-PL} benchmark for Polish.

\subsection{Datasets}
\label{sec:datasets}

\vspace*{-2mm}
\begin{table}[H]
\renewcommand\tabcolsep{4pt}
\setlength\aboverulesep{-1pt}
\setlength\belowrulesep{1pt}
\small
  \centering
 \begin{tabular}{l|c|c|c}
 \toprule
& \multicolumn{2}{c|}{\textbf{NKJP1M}} & {\textbf{PDB-UD}} \\ \midrule
\textsl{POS} & \multicolumn{1}{c|} {Morfeusz} & {Morfeusz / UD} & {\small UD}  \\ 
\textsl{DEP} & \multicolumn{1}{c|} {\small n/a} & {n/a} & {\small UD} \\
\textsl{Format} & \multicolumn{1}{c|} {\small TEI / DAG} & {{\small  CoNLL-X / -U}} & {\small CoNLL-U}  \\ 
\midrule
\# \textsl{tokens} & \multicolumn{2}{c|} {\small 1.2M} & {\small 350K}  \\ 
\# \textsl{sentences} & \multicolumn{2}{c|} {\small 85.7K}  & {\small 22K}  \\ 
\textit{Avg. t/s} & \multicolumn{2}{c|} {\small 14.2} & {\small 15.8}  \\ 
\midrule
\multicolumn{4}{c}{\textbf{NLPre-PL}}\\
\midrule
\textsl{Split} & \multicolumn{1}{c|} {\textsl{byName}} & {\textsl{byType}} &{\textsl{original}} \\ 
\midrule
%\# \textsl{train} & \multicolumn{1}{c|} {69K} &{\small 68K} &{\small 17.7K} \\ 
%\# \textsl{dev} & \multicolumn{1}{c|} {\small 7.7K} &{\small 7.7K} &{\small 2.2K} \\ 
%\# \textsl{test} & \multicolumn{1}{c|} {\small 8.6K} &{\small 9K} &{\small 2.2K} \\
\# \textsl{train} & \multicolumn{1}{c|} {984K} &{\small 978K} &{\small 282K} \\ 
\# \textsl{dev} & \multicolumn{1}{c|} {\small 110K} &{\small 112K} &{\small 35K} \\ 
\# \textsl{test} & \multicolumn{1}{c|} {\small 122K} &{\small 125K} &{\small 34K} \\ 
\bottomrule
  \end{tabular}
  \caption{ \small Summary of source datasets (NKJP1M and PDB-UD) and NLPre-PL Datasets (in tokens). Explanations: \textsl{POS} -- the part-of-speech tagset; \textsl{DEP} -- the dependency schema; \textsl{Avg. t/s} -- the average number of tokens per sentence. }
  \label{tab:stat_nlpre-pl}
  \vspace{-0.2cm}
\end{table}

\paragraph{NKJP1M} \citelanguageresource{NKJP1M}
The NKJP1M subcorpus %\footnote{\href{http://nkjp.pl/index.php?page=14&lang=1}{http://nkjp.pl/index.php?page=14\&lang=1}} % (CC-BY 4.0)} 
of the Polish National Corpus \cite{prz:etal:11:ed}  is manually annotated according to the NKJP tagset \cite{sza:prz:11} and afterwards modified in line with the Morfeusz tagset \cite{woli:19:wuw}. This balanced subset of thematic- and genre-diverse texts and transcriptions  %balanced in terms of gender, age and origin of speakers.  
is used to train Polish POS taggers. NKJP1M %\footnote{The up-to-date version of NKJP1M has been made available to us courtesy of its creators.} 
is maintained in two formats: TEI\footnote{\href{http://nlp.ipipan.waw.pl/TEI4NKJP/}{http://nlp.ipipan.waw.pl/TEI4NKJP}.} %\cite{prze:09c} 
and DAG.\footnote{\href{https://github.com/kawu/concraft-pl\#data-format}{https://github.com/kawu/concraft-pl\#data-format}} 
These two formats are accepted by older NLPre tools but not modern ones. We thus convert NKJP1M to the CoNLL-X format \cite{buchholz-marsi-2006-conll} preserving the original segmentation, POS tags and morphological features (i.e. the Morfeusz tagset), and to the CoNLL-U format\footnote{\href{https://universaldependencies.org/format.html}{https://universaldependencies.org/format.html}} with UD tags, %\footnote{\href{https://universaldependencies.org/u/pos}{https://universaldependencies.org/u/pos}} (\textsl{UPOS}), 
Morfeusz tags (\textsl{XPOS}) and UD morphological features. %\footnote{\href{https://universaldependencies.org/u/feat}{https://universaldependencies.org/u/feat}} (\textsl{UFeats}). 

Since there is no generally accepted split of NKJP1M into training, development and testing subsets, we uniformly divide NKJP1M in all formats (i.e. DAG, TEI, CoNLL-X and CoNLL-U) pursuant to the formulated splitting heuristics. Each document in the subcorpus contains multiple paragraphs of continuous textual data. To avoid possible information leakage, we treat each such paragraph as an indivisible unit. To ensure that the~subsets include paragraphs of varying length, we investigate the distribution over the number of segments in each paragraph. Since it is akin to Gaussian distribution, we decide to not exclude any data, and we divide the paragraphs into K = 10 buckets of roughly similar size and then sample from them with respective ratios of 0.8:0.1:0.1 (corresponding to train, dev, and test subsets). 
This data selection technique assures similar distribution of segments number per paragraph in three subsets, hereafter \textit{byName}. 
For creating our second split, hereafter \textit{byType}, we consider the type of document a paragraph belongs to. We first group paragraphs into categories equal to the document types, and then we repeat the above-mentioned procedure per category 
(see the summary of NKJP1M and data splits in Table~\ref{tab:stat_nlpre-pl}). % and splitting heuristic details in Appendix~\ref{apx:datasets}).

%We provide more information about this methodology in 
%Details on the splitting heuristic are in Appendix~\ref{apx:datasets}.)

%This dataset we call \textit{fairByType} (shortly \textit{byType}).
%The statistical summary of NKJP1M and the proposed data splits is in Table~\ref{tab:stat_nlpre-pl}.
%We provide more information about this methodology in 
%Details on the splitting heuristic are in Appendix~\ref{apx:datasets}.

\vspace*{-5mm}
\paragraph{PDB-UD} \citelanguageresource{PDBUDdataset} Polish Dependency Bank %\footnote{\href{https://github.com/UniversalDependencies/UD_Polish-PDB}{github.com/UniversalDependencies/UD\_Polish-PDB}} 
%\footnote{\href{http://git.nlp.ipipan.waw.pl/alina/PDBUD}{http://git.nlp.ipipan.waw.pl/alina/PDBUD} (It is a~repository with the updated version of Polish-PDB treebank from \href{https://github.com/UniversalDependencies/UD_Polish-PDB}{https://github.com/ UniversalDependencies/UD\_Polish-PDB} (CC BY-NC-SA 4.0 license) }  
%\cite{wroblewska-2018-extended} 
is the largest collection of Polish sentences manually annotated with dependency trees and afterwards converted into UD representations in line with the UD annotation schema \cite{de-marneffe-etal-2021-universal}. PDB-UD slightly correlates with NKJP1M, i.e., a~subset of the PDB-UD sentences comes from NKJP1M, and the language-specific tags (\textsl{XPOS}) in PDB-UD match the Morfeusz tagset. PDB-UD is typically used to train NLPre systems for Polish. In NLPre-PL, we use the original PDB-UD data without any modifications and its standard split (see the statistical summary of PDB-UD in Table~\ref{tab:stat_nlpre-pl}).

\subsection{Tasks}
The complete set of NLPre tasks was originally curated for evaluating language systems in the CoNLL shared task 2018 \cite{zeman-etal-2018-conll}. These tasks mainly focus on preliminary text processing, such as tokenisation or divulging morphosyntactic features. We follow the CoNLL task choice and include all these tasks in NLPre-PL.

\vspace*{-3mm}
\paragraph{Segmentation}
A segmentation task consists in splitting texts into sentences (\textsl{Sentences}), orthographic tokens (\textsl{Tokens}), and syntactic words (\textsl{Words}), the latter being the basic units of morphosyntactic analysis. Segmentation is not a trivial task. In some languages, an orthographic token may be recognised as a \textit{multi-word token} (\textit{multi-word} for short) combining multiple syntactic words, e.g. %in German, the token \textit{im} (Eng. \textit{in}) is a combination of the adposition \textit{in} and the determiner \textit{dem}; 
in Polish, the token \textit{spalibyśmy} (Eng. \textit{we would sleep}) consists of the past participle \textit{spali} (Eng. \textit{slept}), the conditional marker \textit{by} (Eng. \textit{would}) and the mobile inflection \textit{śmy}. 
Since the consistent model of segmentation into words and sentences was used in NKJP1M and PDB-UD, we maintain this data segmentation in NLPre-PL. 
It is also worth mentioning that the CoNLL format (but not TEI and DAG) allows for annotating orthographic tokens; thus, they are included in the NLPre-PL benchmark. 

\vspace*{-3mm}
\paragraph{Tagging}
A tagging task is the process of identifying parts of speech (i.e. POS tagging) and possibly morphological features (i.e. morphological analysis) of words. It follows a predefined POS tagset. 
As mentioned in Section \ref{sec:datasets}, two tagsets are used in the NLPre-PL datasets: Morfeusz and UD. %Although the two tagsets differ, a mapping between them is possible. %, e.g. the Morfeusz tag \texttt{subst:sg:inst:f} corresponds to \textsl{UPOS} (UD part of speech) \texttt{NOUN} and \textsl{UFeats} (UD morphological features) \texttt{Case=Ins|Gender=Fem|Number=Sing}. If the architecture of morphological analyzers allows it, we test them on the NLPre-PL datasets annotated in accordance with both tagsets.
 %ambiguity of words?

\vspace*{-3mm}
\paragraph{Lemmatisation}
Lemmatisation involves predicting canonical forms of syntactic words. Canonical forms are conventionally established identifiers of lexemes (i.e. sets of inflectionally related syntactic words). Since Polish is a fusional language with a~large number of inflected words, lemmatisation is an important task, albeit not trivial, e.g. the lemma of \textit{kluczy} can be either the infinitive \textsc{kluczyć} (Eng. \textit{to weave}) or the noun \textsc{klucz} (Eng. \textit{a key}).

\vspace*{-3mm}
\paragraph{Dependency parsing}
Dependency parsing is the process of automatically predicting the syntactic structure of an input sentence. A dependency structure is a labelled directed tree with nodes corresponding to syntactic words and edges between these words specifying dependency relations.

\setcounter{table}{1}
\begin{table*}[ht!]
\renewcommand\tabcolsep{6pt}
\setlength\aboverulesep{-1pt}
\setlength\belowrulesep{1pt}
 \small
  \centering

 \begin{tabular}{lc||ccc|ccccc||cc}

\textbf{Model / Task} &\rot{\textbf{Average}} & \rot{\textbf{Tokens}}  & \rot{\textbf{Sentences}} & \rot{\textbf{Words}}  & \rot{\textbf{UPOS}} & \rot{\textbf{XPOS}}
& \rot{\textbf{UFeats}} & \rot{\textbf{AllTags}} & \rot{\textbf{Lemmas}} & \rot{\textbf{ Tok/s CPU}} & \rot{\textbf{ Tok/s GPU}}\\
\midrule
\textsl{concraft}  & 91.61 &  98.56  &  71.33  &  99.64  & 95.88 & 90.04 & 90.59 & 90.04 & 96.79 & %3504
% 980.0 & --\\
111 & -- \\
\midrule
\textsl{udpipe} + \textsl{fT}  & 94.43 &  99.75  &  90.51  &  \textbf{99.73}  & 97.36 & 90.64 & 90.97 & 90.64 & 95.86 & %\textbf{43.6} & 45.2\\
2365 & 2181 \\
\midrule
\midrule
\textsl{combo}  + \textsl{fT}  & 95.75 &  99.12  &  \textbf{93.33}  &  99.04  & 97.25 & 93.82 & 93.61 & 92.98 & 96.90 & %468.9 & 145.8\\
458 & 822 \\
%\midrule
\textsl{combo} + \textsl{H} & \textbf{96.67} &  99.12  &  \textbf{93.33}  &  99.04  & \textbf{97.80} & \textbf{95.66} & \textbf{95.75} & \textbf{95.20} & \textbf{97.42} & %827.8 & 166.6\\
241 & 722 \\
\midrule
\textsl{stanza} + \textsl{fT} & 95.89 &  \textbf{99.76}  &  92.70  &  99.45  & 97.43 & 93.57 & 93.90 & 93.36 & 96.94 & %163.1 & 52.1\\
933 & 2379 \\
\midrule
\textsl{spacy}  + \textsl{pl}  & 75.38 &  99.56  &  61.85  &  98.46  & 96.30 & 90.97 & 31.03 & 30.14 & 94.77 &
% 62.6 & \textbf{26.9}\\
\textbf{3252} & \textbf{8407} \\
\textsl{spacy}  + \textsl{fT}  & 75.15 &  99.56  &  61.85  &  98.46  & 95.89 & 89.93 & 31.03 & 30.08 & 94.43 & 
%61.9 & 27.1\\
3134 & 8063 \\
\textsl{spacy} + \textsl{P}& 76.12 & 99.56 & 61.85 & 98.46 & 97.02 & 94.60 & 31.03 & 30.46 & 95.98 & 
%160.9 & 42.3\\
1571 & 5367 \\
\midrule
\textsl{trankit} + \textsl{R}  & 92.59 &  98.37  &  89.39  &  97.84  & 95.36 & 89.74 & 90.05 & 88.73 & 91.19 & %480.3 & 192.3\\
287 & 541 \\
\bottomrule
 \end{tabular}
\caption{\small Results (F1 scores) and inference time (the number of tokens processed per second) of benchmarking the selected NLPre systems on the~Morfeusz tagset averaged by the datasets (\textsl{byName} and \textsl{byType}). The systems are grouped into non-neural and neural by a double horizontal line. Embeddings used in the models are: \textsl{R} -- xlm-RoBERTa-base, \textsl{fT} -- fastText, \textsl{P} -- Polbert, \textsl{pl} -- pl-core-news-lg, \textsl{H} -- HerBERT.}
\label{tab:avgMorfeusz}
\vspace{-0.3cm}
\end{table*}

\begin{table*}[ht!]
\renewcommand\tabcolsep{6pt}
\setlength\aboverulesep{-1pt}
\setlength\belowrulesep{1pt}
 \small
  \centering
 \begin{tabular}{lc||ccc|ccccc||cc}

\textbf{Model / Task} &\rot{\textbf{Average}} & \rot{\textbf{Tokens}}  & \rot{\textbf{Sentences}} & \rot{\textbf{Words}}  & \rot{\textbf{UPOS}} & \rot{\textbf{XPOS}}
& \rot{\textbf{UFeats}} & \rot{\textbf{AllTags}} & \rot{\textbf{Lemmas}} & \rot{\textbf{ Tok/s CPU}} & \rot{\textbf{ Tok/s GPU}}\\
\midrule
\textsl{udpipe} + \textsl{fT}  & 92.30 & \textbf{99.79} & 92.44 & \textbf{99.78} & 97.33 & 89.97 & 90.37 & 89.35 & 95.23 & %51.0 & 51.5\\
1977 & 1848 \\
\midrule
\midrule
\textsl{combo} + \textsl{fT}  & 94.04 & 99.18 & \textbf{94.29} & 98.77 & 96.64 & 93.30 & 93.48 & 91.97 & 96.53 & %415.7 & 135.2\\
 471 & 844 \\
 \textsl{combo} + \textsl{H}  & \textbf{95.51} & 99.21 & \textbf{94.29} & 98.77 & \textbf{97.57} & \textbf{95.33} & \textbf{95.61} & \textbf{94.54} & \textbf{97.13} & %826.8 & 152.6\\
 254 & 733 \\
\midrule
 \textsl{stanza} + \textsl{fT}& 94.25 & 99.77 & 93.92 & 99.43 & 97.33 & 92.88 & 92.90 & 91.63 & 96.60 & 
 %178.6 & 48.2\\
 910 & 2262 \\
\midrule
\textsl{spacy} + \textsl{pl}  & 88.39 & 99.58 & 65.05 & 98.47 & 96.36 & 90.95 & 91.22 & 89.65 & 93.62 & 
%40.1 & 39.9\\
\textbf{2495} & \textbf{5403} \\
\textsl{spacy} + \textsl{fT}  & 87.68 & 99.58 & 65.05 & 98.47 & 95.79 & 89.77 & 90.05 & 88.37 & 93.37 & 
2484 & 4533 \\
 \textsl{spacy} + \textsl{P}  & 90.70 & 99.58 & 65.05 & 98.47 & 97.26 & 94.68 & 94.84 & 94.09 & 94.89 & 
 %173.5 & 48.4\\
 1376 & 4207 \\
\midrule
 \textsl{trankit} + \textsl{R}& 92.91 & 98.88 & 92.44 & 98.52 & 96.50 & 91.74 & 91.91 & 90.21 & 90.47 & 
 %439.8 &172.6\\
 319 & 593 \\
\bottomrule
 \end{tabular}
\caption{\small Results (F1 scores) and inference time (tokens per second) of benchmarking the selected NLPre systems on the UD tagset averaged by the datasets (\textsl{byName}, \textsl{byType}, and \textsl{PDB-UD}). The systems are grouped into non-neural and neural by a double horizontal line (Concraft is not included because it does not allow data in the UD tagset) Embeddings used in the models are: \textsl{R} -- xlm-RoBERTa-base, \textsl{fT} -- fastText, \textsl{P} -- Polbert, \textsl{pl} -- pl-core-news-lg, \textsl{H} -- HerBERT.}
\label{tab:avgUD}
\vspace{-0.3cm}
\end{table*}

\begin{table*}[ht!]
\renewcommand\tabcolsep{8pt}
\setlength\aboverulesep{-1pt}
\setlength\belowrulesep{1pt}
\small
  \centering
 \begin{tabular}{lc||c|c|c|c|c|c|c||c}
\textbf{Task / model} & \textsl{\rot{{udpipe + fT}}}  & \textsl{\rot{{combo + fT}}} & \textsl{\rot{{combo + H}}}  & \textsl{\rot{{stanza + fT}}} & \textsl{\rot{{spacy + fT}}} & \textsl{\rot{{spacy + pl}}} & \textsl{\rot{{spacy + P}}} & \textsl{\rot{{trankit + R}}} & \color{gray}{\textsl{\rot{{GPT-3.5}}}} \\
\midrule
\textbf{Avg. F1 on PDB-UD} & 88.16 & 90.46 & 93.37 & 92.10 & 83.03 & 84.21 & 87.98 & \textbf{94.03}  & \color{gray}{50.95}\\
\midrule
\textbf{Tokens} & 99.86 & 99.35 & 99.40 & 99.86 & 99.65 & 99.65 & 99.65 & \textbf{99.90} & \color{gray}{98.08}\\
\textbf{Sentences} & 95.90 & 96.22 & 96.22 & 96.83 & 71.46 & 71.46 & 71.46 & \textbf{98.51}  & \color{gray}{89.81}\\
\textbf{Words} & 99.84 & 98.22 & 98.22 & 99.42 & 98.51 & 98.51 & 98.51 & \textbf{99.89}  & \color{gray}{96.96} \\
\midrule
\textbf{UPOS} & 97.28 & 95.34 & 97.31 & 97.64 & 95.62 & 96.49 & 97.54 & \textbf{99.07} & \color{gray}{64.07} \\
\textbf{XPOS} & 88.57 & 92.03 & 94.92 & 93.17 & 88.57 & 90.14 & 94.35 & \textbf{96.18}  & \color{gray}{41.32}\\
\textbf{UFeats} & 89.07 & 92.21 & 95.23 & 93.22 & 88.79 & 90.42 & 94.52 & \textbf{96.34} & \color{gray}{41.88}\\
\textbf{AllTags} & 88.02 & 90.41 & 94.29 & 92.15 & 87.00 & 88.71 & 93.85 & \textbf{95.57}  &\color{gray}{35.65} \\
\textbf{Lemmas} & 94.29 & 95.37 & \textbf{96.38} & 95.77 & 91.72 & 91.69 & 93.77 & 88.98 & \color{gray}{64.77}\\
\midrule
\textbf{UAS} & 86.68 & 88.49 & 91.31 & 91.09 & 80.91 & 82.15 & 88.08 & \textbf{95.79} & \color{gray}{35.57} \\
\textbf{LAS} & 83.01 & 86.19 & 89.98 & 88.83 & 72.24 & 73.60 & 80.33 & \textbf{94.24}  & \color{gray}{26.58}\\
\textbf{CLAS} & 79.53 & 84.14 & 89.03 & 86.90 & 73.72 & 75.71 & 80.50 & \textbf{93.00} & \color{gray}{29.06}\\
\textbf{MLAS} & 69.53 & 76.64 & 84.77 & 79.90 & 63.57 & 66.90 & 75.75 & \textbf{87.79} & \color{gray}{11.81}\\
\textbf{BLEX} & 74.49 & 81.34 & \textbf{86.77} & 82.53 & 67.64 & 69.29 & 75.48 & 77.18 &\color{gray}{26.86} \\
\midrule
\midrule
\textbf{Avg. F1 on NKJP1M} & 94.37 & 95.84 & \textbf{96.59} & 95.33 & 90.01 & 90.49 & 92.06 & 92.36 & \color{gray}{NA} \\

\bottomrule
 \end{tabular}
\caption{\small Results of benchmarking the selected NLPre systems on the smaller PDB-UD dataset. The last row with the mean F1 scores of the models trained on larger NKJP1M data is for reference. Embeddings used in the models are: \textsl{R} -- xlm-RoBERTa-base, \textsl{fT} -- fastText, \textsl{P} -- Polbert, \textsl{pl} -- pl-core-news-lg, \textsl{H} -- HerBERT. The results of GPT-3.5 are greyed out due to their exclusion from display on the leaderboard.}
\label{tab:ud-pdb-results-pl}
\vspace{-0.2cm}
\end{table*}
\section{Evaluation}
\label{sec:evaluation}

\subsection{Evaluation methodology}
To maintain the de facto standard to NLPre evaluation, we apply the evaluation measures defined for the CoNLL 2018 shared task %\cite{zeman-etal-2018-conll} 
and implemented in the official evaluation script.\footnote{\href{https://universaldependencies.org/conll18/conll18_ud_eval.py}{\scriptsize{https://universaldependencies.org/conll18/conll18\_ud\_eval.py}}}  In particular, we focus on F1 and \textit{AlignedAccuracy}, which is similar to F1 but does not consider possible misalignments in tokens, words, or sentences.

In our evaluation process, we follow default training procedures suggested by the authors of the evaluated systems, i.e. we do not conduct any optimal hyperparameter search in favour of leaving the recommended model configuration as-is. We also do not further fine-tune selected models.

% Research in the field of natural language preprocessing has been conducted for a long time and during this time a detailed methodology of evaluating the considered tasks has been established. Measures for the evaluation of individual tasks were defined and refined mainly in numerous shared tasks to best serve result interpretation and model comparison. We therefore adopt them to evaluate the tools tested on our benchmark.

%For the purpose of evaluation of the preprocessing models for Polish language with NLPre-PL benchmark, we select well-rooted rule-based disambiguation methods and more modern systems based on neural network architectures to assemble an informative and thorough comparison of different possible approaches. We present a comprehensive overview of selected systems below.
%Based on the NLPre-PL benchmark, we evaluate morphosyntactic analyses with three kinds of architectures: (1) pipelines of separate tools (i.e. Concraft and UDPipe), (2) systems integrating separate models for preprocessing tasks (i.e. SpaCy, Stanza, and Trankit), and (3) end-to-end systems with one model for all preprocessing tasks (i.e. COMBO).

\subsection{Evaluated systems}
Based on the NLPre-PL benchmark, we evaluate both well-rooted rule-based disambiguation methods and modern systems based on neural network architectures to enable an informative and thorough comparison of different approaches. We use the most up-to-date versions of available tools at the time of conducting experiments: 
(1) pipelines of separate tools (Concraft-pl, UDPipe), % \citep{udpipe:2017}, 
(2) systems integrating separate models for NLPre tasks (spaCy, Stanza, Trankit), (3) end-to-end systems with a model for all NLPre tasks (COMBO), and large language model GPT-3.5. %\cite{ouyang2022training} These systems are detailed in Appendix~\ref{apx:systems}. 

\vspace*{-3mm}
\paragraph{Concraft-pl} \cite{waszczuk2012harnessing,waszczuk2018morphosyntactic} \label{app:concraft}\footnote{Polish is a fusional language for which a two-stage tagging procedure is typically applied: first, a~rule-based morphological analyser outputs all morphological interpretations of individual tokens, and then a~tagging disambiguator selects the most likely one for each token. The tools implemented in accordance with this procedure are still imminent.} is a system for joint morphosyntactic disambiguation and segmentation.\footnote{\href{https://github.com/kawu/concraft-pl}{https://github.com/kawu/concraft-pl} (v2.0)} It uses Morfeusz morphological analyser \cite{wol:14,kie:wol:17:morf} to extract morphological and segmentation equivocates and then disambiguates them using the conditional random fields model. We train the Concraft-pl models with default parameters.

\vspace*{-3mm}
\paragraph{UDPipe} \cite{udpipe:2017} \label{app:udpipe} is a~language-agnostic trainable NLPre pipeline.\footnote{\href{https://ufal.mff.cuni.cz/udpipe}{https://ufal.mff.cuni.cz/udpipe} (v1)} Depending on the task, it uses recurrent neural networks \cite{GRAVES2005602} in segmentation and tokenization, the average perceptron in tagging and lemmatization, a~rule-based approach in multi-word splitting, and a transition-based neural dependency parser. We train the UDPipe models with the default parameters. The dependency parser is trained with the Polish \textit{fastText} embeddings \citelanguageresource{fasttext}.%\cite{bojanowski-etal-2017-enriching}.

\vspace*{-3mm}
\paragraph{SpaCy}  
\cite{spacy3} \label{app:spacy} is an NLP Python library shipped with pretrained pipelines and word vectors for multiple languages.\footnote{\href{https://github.com/explosion/spaCy}{https://github.com/explosion/spaCy} (v3.4.1)} It also supports training the models for tagging and parsing, inter alia. We use spaCy to train pipelines for morphosyntactic analysis with: feed-forward network-based text encoders with static embeddings (\textit{fastText} and \textit{pl-core-news-lg}) or transformer-based encoders  %(\textit{dkleczek/bert-base-polish-cased-v1})
with the Polbert embeddings \citelanguageresource{polbert}, %\footnote{\href{https://huggingface.co/dkleczek/bert-base-polish-cased-v1}{https://huggingface.co/dkleczek/bert-base-polish-cased-v1}}
 taggers (linear layers with softmax activation on top of the encoders), and transition-based parsers.

\vspace*{-3mm}
\paragraph{Stanza}\cite{qi-etal-2020-stanza} \label{app:stanza} is a language-agnostic, fully neural toolkit offering a~modular pipeline for tokenization, multi-word token expansion, lemmatization, tagging, and dependency parsing.\footnote{\href{https://github.com/stanfordnlp/stanza}{https://github.com/stanfordnlp/stanza} (v1.4.0)} It mainly uses recurrent neural networks \cite{GRAVES2005602} as a base architecture and external word embeddings (\textit{fastText}). Each module reuses the basic architecture.

\vspace*{-3mm}
\paragraph{Trankit} \cite{nguyen2021trankit}\label{app:trankit} uses a~multilingual pre-trained transformer-based language model, XLM-Roberta \citelanguageresource{xmlroberta} %\footnote{\href{https://huggingface.co/xlm-roberta-base}{https://huggingface.co/xlm-roberta-base}} \cite{conneau-etal-2020-unsupervised}, 
as the text encoder which is then shared across pipelines for different languages.\footnote{\href{https://github.com/nlp-uoregon/trankit}{https://github.com/nlp-uoregon/trankit} (v1.1.1)} The resulting model is jointly trained on 90 UD treebanks with a~separate adapter \cite{pfeiffer-etal-2020-adapterhub,pfeiffer-etal-2020-mad} for each treebank. Trankit uses a~wordpiece-based splitter to exploit contextual information. %Beside standard preprocessing tasks it also implements a name entity recognizer. 

\vspace*{-3mm}
\paragraph{COMBO} \cite{rybak-wroblewska-2018-semi,klimaszewski-wroblewska-2021-combo-state}
%\lp{
\label{app:combo} is a fully neural language-independent NLPre system\footnote{\href{https://gitlab.clarin-pl.eu/syntactic-tools/combo}{https://gitlab.clarin-pl.eu/syntactic-tools/combo} (v1.0.5)} integrated with the LAMBO tokeniser \cite{lambo}. It is an end-to-end system with jointly trained mo\-du\-les for tagging, parsing, and lemmatisation. We train the COMBO models with the pre-trained word embeddings -- \textit{fastText} and \textit{HerBERT} \citelanguageresource{herbert}. % \cite{mroczkowski-etal-2021-herbert}. %Since more than one feature per word can be provided representation is averaged. Encoder - BLSTM. Tagger (XPOS, UPOS, UFEATS -  FC layers. Lemma - CNN. Parser (dependency trees and head) - ... #draft informacji %}

\vspace*{-3mm}
\paragraph{GPT-3.5} \cite{brown:2020} is a large language model, notable for its outstanding performance in NLU tasks. It is a fined-tuned version of the GPT-3 model. GPT-3.5's architecture is based on a transformer neural network with 12 stacks of decoders blocks with multi-head attention blocks.

%\vspace{2mm}
For segmentation tasks, we train modules integrated with the tested NLPre systems. The only aberration is in %the case of 
spaCy, %where we obtained poor segmentation results of the dependency module.\footnote{Dependency parsing module is responsible for sentence segmentation in the newest implementation of spaCy.} and decided to use an~out-of-the-box sentenciser available in spaCy.
where poor segmentation results of the dependency module\footnote{Dependency parsing module is responsible for sentence segmentation in the spaCy implementation.} forced us to use an~out-of-the-box sentenciser available in spaCy.

For each model, we initialise training with possibly the most prominent and congruent embedding model available. Virtually all models are capable of fully capitalising from that addition, apart from Concraft and UDPipe. The first does not use embeddings at all, and the latter uses them only for dependency parsing training. If embeddings based on BERT architecture are feasible to use, we select their \textit{base} versions. This ensures fairness of comparison between NLPre systems, as not all of them support BERT-\textit{large} embeddings.

\subsection{Results}
\paragraph{Impact of system architecture}
We assess the quality of the selected NLPre systems contingent on the NLPre-PL benchmark. %We focus on comparing segmentation tools, POS taggers, morphological analysers and lemmatizers. 
In Polish (and most other languages), non-neural NLPre tools are currently not widely developed. We evaluate two of them: Concraft %\footnote{Concraft accepts processing data only in the Morfeusz tagset, it is thus not used in the UD evaluation experiment.} 
and UDPipe. Although they do not use neural network algorithms to train models, their quality does not significantly differ from the best tested neural systems, especially in terms of segmentation, which UDPipe performs best (\textsl{Words}) or second-best (\textsl{Sentences}) (see Tables \ref{tab:avgMorfeusz} and \ref{tab:avgUD}). We cannot unequivocally say that the system architecture has a decisive influence on the results, as spaCy models, even transformer-based, output the lowest~quality.   

\vspace*{-2mm}
\paragraph{Impact of tagset selection} We compare systems trained and tested on data adjusted to two tagsets -- the Morfeusz tagset (see Table \ref{tab:avgMorfeusz}) and the UD tagset (see Table \ref{tab:avgUD}). The average scores indicate that only COMBO performs better on Morfeusz-annotated data than on UD data. The performance of Trankit, UDPipe, and Stanza slightly decreases on Morfeusz data. Notably, all spaCy models trained on this dataset record a significant quality drop mainly due to poorly performed morphological analysis, i.e. \textsl{UFeats} values (and thus also the low \textsl{AllTags} values, i.e., matching between \textsl{UPOS}, \textsl{XPOS}, and \textsl{UFeats}). Regarding segmentation, \textsl{UPOS} and \textsl{XPOS} tagging, and lemmatisation, the tagset selection does not negatively affect the results, and the systems perform comparably.

\vspace*{-2mm}
\paragraph{Impact of the size of training data}
Intuitively, the size of the training data affects the prediction quality. Considering the data size factor, we compare the average F1 scores of the NLPre systems trained on NKJP1M (see the last row in Table \ref{tab:ud-pdb-results-pl}) and on PDB-UD (see Table \ref{tab:ud-pdb-results-pl}), which is two orders of magnitude smaller. The results confirm our intuitive assumptions -- there is a difference of 6.21 between the mean F1 scores obtained by the systems trained on the smaller PDB-UD %(i.e. the average F1 of 88.16) 
(avg. F1 of 88.16) and those trained on the larger NKJP1M %(i.e. the average F1 of 94.37).
(avg. F1 of 94.37).

When comparing the performance of individual systems on the smaller PDB-UD dataset, Trankit turns out to be the undisputed winner in all tasks except lemmatisation. However, considering the average performance of all tasks, COMBO and Stanza perform the best. 

In alignment with contemporary developments on zero-shot learning, we test the predictive capabilities of GPT-3.5 acquired via the prompting technique \cite{brown:2020}. Despite comprehensive instructions along with the UD tree examples in the prompt, the results are highly unsatisfactory. An error analysis has revealed that 1) the GPT model modifies the input texts (e.g. adds elided words, alters the word's declension and conjugation, leading also to non-existent words); 2) while parsing questions, it answers them or returns information that they cannot be answered; 3) it replaces Polish words with their foreign equivalents; 4) it outputs graphs with cycles, thus not adhering to UD trees. Even for GPTs, achieving UD-compliant morphosyntactic analysis is challenging when they lack access to training examples. GPT-3.5's results are not included in the leaderboard. %highlighting the need for its further investigation, e.g. LLM's fine-tuning.

\vspace*{-2mm}
\paragraph{Impact of split heuristics}
As outlined in Section \ref{sec:datasets}, NKJP1M has no official split into train, dev, and test subsets. Since intuitively, the type of document can affect text processing, we propose two alternative splits, i.e. \textit{byName} and \textit{byType}. %To verify the impact of data splitting, we compare the F1 scores averaged across the tasks and the selected systems trained and tested on the \textit{byName} data -- 90.69 vs. the \textit{byType} data -- 90.56. 
We compare the F1 scores for these two splits to verify this hypothesis. For the \textit{byName} split, the average F1 for tasks and systems is 90.69, and for the \textit{byType} split, it is 90.56.
The difference is negligible, indicating that the document type, and hence the text domain, does not affect the quality of the NLPre tasks. Based on this outcome, we arbitrarily choose the more balanced \textit{byType} split as binding in the final NLPre-PL benchmarking system. The detailed results of all experiments are in Appendix \ref{apx:full_results}.

\vspace*{-2mm}
\paragraph{Inference time}
%Considering only the subset of tasks described in this paragraph, it is noteworthy that the best average F1 results are obtained by COMBO and Stanza, leaving spaCy and Concraft significantly behind. In benchmarking context, it is also valuable to examine at least inference time
In the context of benchmarking, quality is a~fundamental factor. In our case, the best average F1 scores are achieved by COMBO and Stanza, far ahead of spaCy and Concraft. The second crucial issue is the processing time of the evaluated NLPre systems, especially their inference time.\footnote{We share a conviction favoured in the NLP community that the training time is slightly less requisite than the inference time since models are trained only once but then constantly reused for predictions. We thus provide inference times.} % of the tested systems.} 
We calculate the times in which the systems tokenise, tag and lemmatise the input text.\footnote{We run tests uniformly on CPU -- Intel Xeon Platinum 8268 processor (1 node with 12 cores), and GPU -- 2x Tesla V100-SXM2. The machines used to train the models are listed in Appendix \ref{apx:infrastructure}.} The exception is COMBO with the mandatory parsing module that cannot be disabled. Therefore, its calculations include the parsing time as well. The inference time, corresponding to the number of tokens processed per second, is provided in the last two columns of Tables \ref{tab:avgMorfeusz} and \ref{tab:avgUD}. On CPU, the fastest systems are spaCy and UDPipe, and the slowest is Concraft. Other systems process one order of magnitude fewer tokens per second than the top ones. On GPU, spaCy is the undisputed winner, followed by Stanza, UDPipe, COMBO and Trankit.

\begin{figure}[H]%[ht!]
\centering
\includegraphics[width=0.5\textwidth]{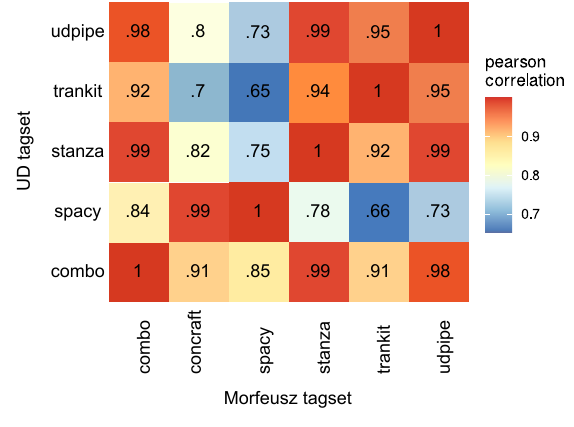}
\caption{\small Pearson correlation coefficients between vectors of F1 scores on \textsl{Tokens}, \textsl{Sentences},
\textsl{Words}, \textsl{UPOS}, \textsl{XPOS}, \textsl{Lemmas} tasks averaged %across 
over datasets (excluding PDB-UD) and embeddings.}
\label{rplot}
\vspace*{-5mm}
\end{figure}

%\vspace*{-3mm}
\paragraph{Correlation analysis} We conduct a~statistical a\-na\-ly\-sis to capture meaningful relations between the performance and the model types, the used embeddings, or the datasets. To check whether the performance of a~given model on a~given tagset allows us to expect similar relationships between the scores on another tagset, we calculate a~correlation matrix of vectors composed of the F1 scores for various tasks, i.e. $\vec{v} =$ [\textsl{Tokens}, \textsl{Sentences}, \textsl{Words}, \textsl{UPOS}, \textsl{XPOS}, \textsl{Lemmas}], averaged over embeddings and datasets (see Figure \ref{rplot}). The vectors are calculated for a pair (\textit{tagset}$_i$, \textit{model}$_j$). To maintain comparability, we exclude PDB-UD from the study as it does not appear in the Morfeusz tagset.

\vspace*{-2mm}

\begin{figure}[H]%[ht!]
\centering
\includegraphics[width=0.45\textwidth]{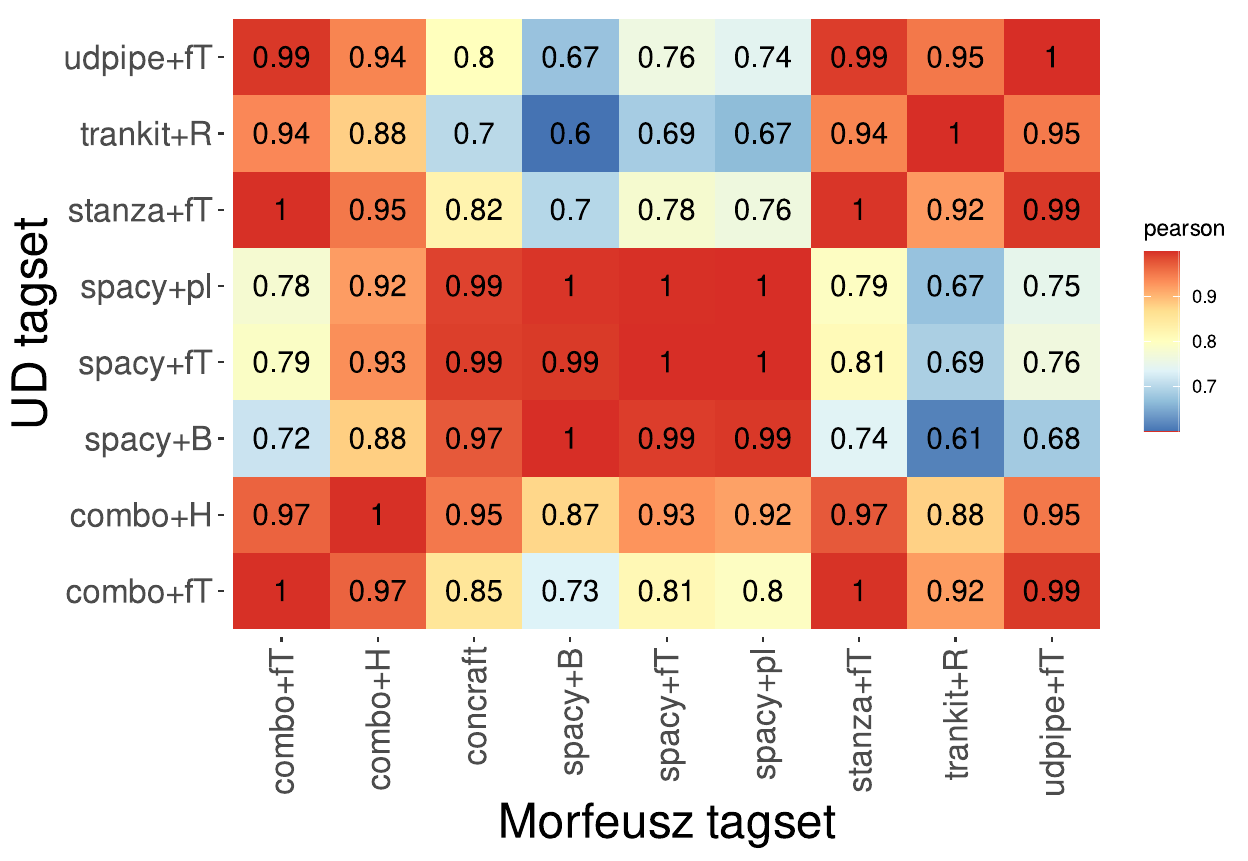}
\caption{\small Pearson correlation coefficients between vectors of F1 scores on \textsl{Tokens}, \textsl{Sentences}, \textsl{Words}, \textsl{UPOS}, \textsl{XPOS}, \textsl{Lemmas} tasks averaged %across 
over datasets (excluding PDB-UD).}
\label{rplot01}
%\vspace*{-5mm}
\end{figure}

Pearson's correlation $r$ suggests that the results are linearly proportional for the same models and different tagsets, which we conclude from the values close to 1 at the intersection of (\textit{model}$_i$, \textit{tagset}$_{\textsc{ud}}$) and (\textit{model}$_i$, \textit{tagset}$_{\textsc{nkjp}}$). Even though correlation coefficients are generally high (i.e. $r \in [0.90, 0.99]$) for most pairs (\textit{model}$_i$, \textit{tagset}$_{\textsc{ud}}$) and (\textit{model}$_{j}$, \textit{tagset}$_{\textsc{nkjp}}$), there are noticeable lower values for spaCy, i.e. $r \in [0.66, 0.78]$. We hypothesise that this is due to the non-linear rate of changes between the scores, as all Spearman correlation coefficients exceed $0.89$ (i.e. $\rho > 0.89$).

\vspace*{-3mm}
\begin{figure}[H]
\includegraphics[width=0.5\textwidth]{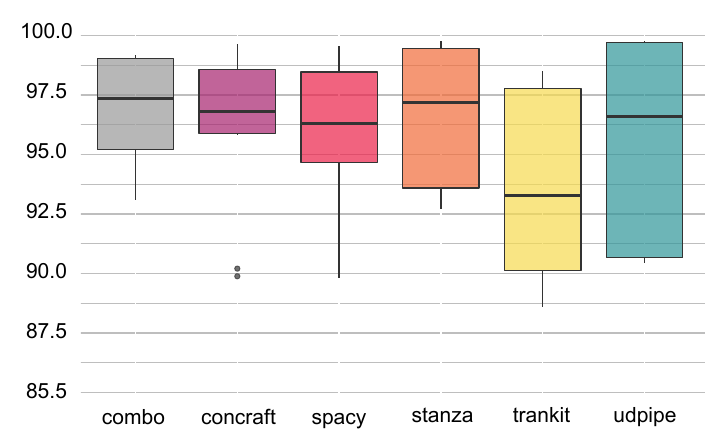}
\caption{\small Dispersion of model performance measured by F1 on the Morfeusz tagset and \textsl{Sentences}, \textsl{Words}, \textsl{UPOS}, \textsl{XPOS}, and \textsl{Lemmas} tasks.}
\label{fig:rplot02}
 \vspace*{-5mm}
\end{figure}

\vspace*{-3mm}
\begin{figure}[H]
\includegraphics[width=0.5\textwidth]{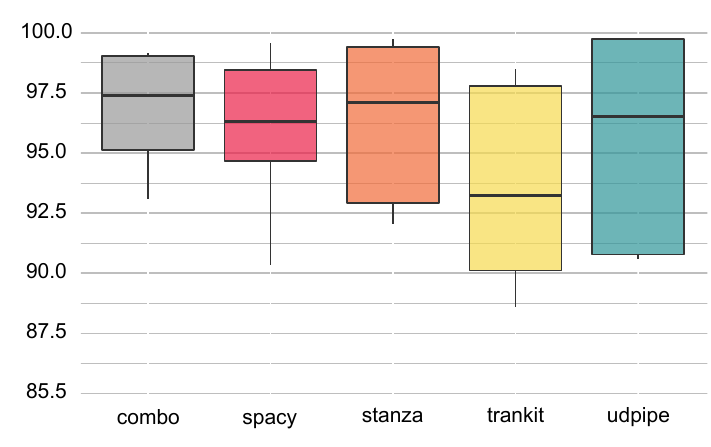}
\caption{\small Dispersion of model performance measured by F1 on the UD tagset and \textsl{Sentences}, \textsl{Words}, \textsl{UPOS}, \textsl{XPOS}, and \textsl{Lemmas} tasks.}
\label{rplot03}
\vspace*{-3mm}
\end{figure}

The results of a more granular analysis of Pearson's $r$ %correlation coefficients 
between vectors of F1 scores
for triples (tagset$_{i}$, model$_{j}$, embeddings$_{k}$), averaged %across 
over datasets, show a strong correlation for the same models, regardless of the tagset and the embedding (see Figure \ref{rplot01}). Hence, if a change in the tagset or embedding causes an increase in one task, a proportional increase in remaining tasks is~expected.

Boxplot charts (see Figures \ref{fig:rplot02} and \ref{rplot03}) determine the stability of the model results for a given tagset regardless of dataset and embedding. One box shows the scattering of F1 scores for \textsl{Tokens}, \textsl{Sentences}, \textsl{Words}, \textsl{UPOS}, \textsl{XPOS}, and \textsl{Lemmas} tasks. The shortest %smallest height of 
COMBO's box indicates a~relatively similar performance of the model across tasks for each triplet (COMBO, \textit{embedding}$_{j}$, \textit{dataset}$_{k}$).

\section{Conclusions}
\label{sec:conclusions}
In this work, we propose a revised approach to NLPre evaluation via benchmarking. This is motivated by the widespread use of the benchmarking technique in other NLP fields on par with the shortcomings of existing NLPre evaluation solutions. 

We implement said NLPre benchmarking approach as the online system that evaluates the submitted outcome of an NLPre system and updates the associated leaderboard with the results after the submitter's approval. The benchmarking system is designed to rank NLPre tools available for a given language in a trustworthy environment.

The endeavour of defining and enhancing the system's capabilities is conducted concurrently with the effort to create the \hbox{NLPre} benchmark for Polish that encompasses numerous factors, such as tasks not required in English or diverse tagsets. The NLPre-PL benchmark %integrated with the benchmarking system 
%The preliminary version of this language-centric benchmarking system has been configured for Polish and integrated with the new NLPre-PL performance benchmark. NLPre-PL 
consists of the predefined NLPre tasks, coupled with two reformulated datasets. %, prepared for prediction assessment. 
The NLPre-PL benchmark, therefore, sets the standard for evaluating the performance of the NLPre tools for Polish,  which represents a~derivative yet important outcome of our research.

In addition to integration into the benchmarking system, NLPre-PL is used to conduct empirical experiments. We perform a robust and extensive comparison of different NLPre methods, including the classical non-neural tools and the modern neural network-based techniques. The results of these experiments on datasets in two tagsets %, i.e. Morfeusz and UD, 
are discussed in detail. The experiments confirm our assumptions that modern architectures obtain better results. 
Because NLP is a discipline undergoing rapid progress, new NLPre solutions, e.g. multilingual or zero-shot, can be expected in the coming years. These new solutions can be easily tested and compared with the tools evaluated so far in our benchmarking system.

Finally, we release the open-source code of the benchmarking system in hopes that this endeavour could be replicated for other languages. To expedite this process, we ensure that the system is fully configurable and language- and tagset-agnostic. The NLPre system, configured for a specified language, can be self-hosted on a chosen server, and the results from the leaderboard are conveniently accessible via an API.  We see a potential future application of our system to the UD repository, where for 141 languages, there are currently 245 treebanks with supposedly discrepant versions of the UD tagset.

\section{Appendices}
\label{sec:appendix}

\subsection{Infrastructure used}
\label{apx:infrastructure}
We train the models using several types of computational nodes at our disposal, including NVIDIA V100 32GB, NVIDIA GeForce RTX 2080 8GB, NVIDIA GeForce 3070 8GB and Intel Xeon E5-2697 processor. Since we do not perform hyperparameter tuning, this should not impact our results.

\subsection{Further results of experiments}
\label{apx:full_results}
Herein, we present a comprehensive depiction of our experimental findings as they are displayed on the NLPre-PL leaderboard.
%the full results of our experiments as they appear on the leaderboard. 

In Table \ref{tab:5FullMorfeusz}, we present the full results of the evaluation of the selected models on the Morfeusz-based datasets \textit{byName} and \textit{byType}. These results are provided for all available tasks that can be performed on the above-mentioned datasets. As NKJP1M datasets contain no syntantic trees, it is thus impossible to test the dependency parsing task that rely on these trees and measure \textsl{UAS}, \textsl{LAS}, \textsl{CLAS}, \textsl{MLAS} and \textsl{BLEX}.

In Table \ref{tab:6UDonTagging}, we present the results of the evaluation of the selected models on the UD-based datasets \textit{byName}, \textit{byType}, and \textit{PDB}. This table contains the results of segmentation, tagging, and lemmatization tasks. Table \ref{tab:7UDonparsing} is a continuation of Table \ref{tab:6UDonTagging} and it contains the results for the same tagset and dataset on the dependency parsing task.

\begin{table*}[p]
%\begin{subtable}{\textheight}
\centering
\small
 \begin{tabular}{lc|c|c|c|c|c|c|c|c|c|c}
{Model / Task} & \rot{\textbf{Dataset}} &\rot{\textbf{Scores}} &\rot{\textbf{Average}} & \rot{\textbf{Tokens}}  & \rot{\textbf{Sentences}} & \rot{\textbf{Words}}  & \rot{\textbf{UPOS}} & \rot{\textbf{XPOS}} & \rot{\textbf{UFeats}} & \rot{\textbf{AllTags}} & \rot{\textbf{Lemmas}} \\
\midrule
\multirow{1}{*}{\begin{tabular}[c]{@{}l@{}}\textsl{combo} \\ + \textsl{H}\end{tabular}} & bN & AA & 97.31 & - & - & - & 98.74 & 96.63 & 96.70 & 96.15 & 98.36\\
& bN & F1 & 96.68 &	99.07 &	93.57 &	99.01 &	97.76 &	95.67 &	95.74 &	95.20 &	97.39\\
& bT & AA & 97.28 &	- &	- &	- &	98.76 &	96.54 &	96.65 &	96.08 &	98.35\\
& bT & F1 & 96.65 &	99.16 &	93.08 &	99.07 &	97.84 &	95.65 &	95.76 &	95.19 &	97.44\\
\midrule
\multirow{1}{*}{\begin{tabular}[c]{@{}l@{}}\textsl{stanza} \\ + \textsl{fT}\end{tabular}} & bN & AA & 95.58 & 	- & 	- & 	- & 	97.97 & 	94.09 & 	94.44 & 	93.89 & 	97.51\\
& bN & F1 & 95.88 &	99.75 &	92.69 &	99.43 &	97.41 &	93.55 &	93.91 &	93.36 &	96.96\\
& bT & AA & 95.55 &	- &	- &	- &	97.97 &	94.10 &	94.38 &	93.85 &	97.43\\
& bT & F1 & 95.89 &	99.77 &	92.70 &	99.46 &	97.45 &	93.59 &	93.88 &	93.35 &	96.9\\
\midrule
\multirow{1}{*}{\begin{tabular}[c]{@{}l@{}}\textsl{combo} \\ + \textsl{fT}\end{tabular}} & bN & AA & 95.87 &	- &	- &	- &	98.15 &	94.81 &	94.60 &	93.97 &	97.80\\
& bN & F1 & 95.78 &	99.07 &	93.57 &	99.01 &	97.18 &	93.87 &	93.67 &	93.04 &	96.84\\
& bT & AA & 95.78 &	- &	- &	- &	98.22 &	94.63 &	94.41 &	93.77 &	97.85\\
& bT & F1 & 95.72 &	99.16 &	93.08 &	99.07 &	97.31 &	93.76 &	93.54 &	92.91 &	96.95\\
\midrule
\multirow{1}{*}{\begin{tabular}[c]{@{}l@{}}\textsl{udpipe} \\ + \textsl{fT}\end{tabular}}& bN & AA & 93.34 &	- &	- &	- &	97.57 &	90.90 &	91.22 &	90.90 &	96.12\\
& bN & F1 & 94.44 &	99.77 &	90.43 &	99.75 &	97.33 &	90.68 &	90.99 &	90.68 &	95.88\\
& bT & AA &93.35 &	- &	- &	- &	97.67 &	90.87 &	91.21 &	90.87 &	96.13\\
& bT & F1 & 94.42 & 99.73 &	90.58 &	99.70 &	97.38 &	90.60 &	90.94 &	90.60 &	95.84\\
\midrule
\multirow{1}{*}{\begin{tabular}[c]{@{}l@{}}\textsl{trankit} \\ + \textsl{R}\end{tabular}} & bN & AA & 93.06 &	- &	- &	- &	97.49 &	91.77 &	92.05 &	90.73 &	93.25\\
& bN & F1 & 92.81 &	98.50 &	90.19 &	97.96 &	95.50 &	89.89 &	90.17 &	88.88 &	91.35\\
& bT & AA & 92.99 &	- &	- &	- &	97.43 &	91.69 &	92.03 &	90.65 &	93.15\\
& bT & F1 & 92.36 &	98.24 &	88.58 &	97.72 &	95.21 &	89.59 &	89.93 &	88.58 &	91.02\\
\midrule
\multirow{1}{*}{\begin{tabular}[c]{@{}l@{}}\textsl{concraft} \\ \end{tabular}} & bN & AA & 93.09 &	- &	- &	- &	96.24 &	90.51 &	91.05 &	90.51 &	97.13\\
& bN & F1 & 91.70 &	98.56 &	71.55 &	99.65 &	95.90 &	90.20 &	90.73 &	90.20 &	96.79\\
& bT & AA & 92.92 &	- & - &	- &	96.22 &	90.22 &	90.79 &	90.22 &	97.15\\
& bT & F1 & 91.52 &	98.55 &	71.10 &	99.62 & 95.86 &	89.88 &	90.45 &	89.88 &	96.79\\
\midrule
\multirow{1}{*}{\begin{tabular}[c]{@{}l@{}}\textsl{spacy} \\ + \textsl{P}\end{tabular}} & bN & AA & 70.94  & - & -  &	-  & 98.54  & 96.12  &	31.54  & 30.96  & 97.52\\
& bN & F1 & 76.23  & 99.56  & 62.64  & 98.45  &	97.01  & 94.64  &	31.05  & 30.48  & 96.01\\
& bT & AA & 70.88  & -  & -  &	-  & 98.55  & 96.03  &	31.49  &	30.91  & 97.44\\
& bT & F1 & 76.00  & 99.56  & 61.06  & 98.46  & 97.03  & 94.55  & 31.00  & 30.43  & 95.94 \\
\midrule
\multirow{1}{*}{\begin{tabular}[c]{@{}l@{}}\textsl{spacy} \\ + \textsl{pl}\end{tabular}} & bN & AA & 69.77 &	- &	- &	- &	97.86 &	92.47 &	31.54 &	30.68 &	96.30\\
& bN & F1 & 75.51 &	99.56 &	62.64 &	98.45 &	96.34 &	91.04 &	31.05 &	30.21 &	94.81\\
& bT & AA & 69.66 &	- &	- &	- &	97.77 &	92.31 &	31.49 &	30.54 &	96.21\\
& bT & F1 & 75.25 &	99.56 &	61.06 &	98.46 &	96.26 &	90.89 &	31.00 &	30.07 &	94.73\\
\midrule
\multirow{1}{*}{\begin{tabular}[c]{@{}l@{}}\textsl{spacy} \\ + \textsl{fT}\end{tabular}} & bN & AA & 69.39 &	- &	- &	- &	97.42 &	91.48 &	31.54 &	30.61 &	95.89\\
& bN & F1 & 75.28 &	99.56 &	62.64 &	98.45 &	95.92 &	90.06 &	31.05 &	30.13 &	94.40\\
& bT & AA & 69.29 &	- &	- &	- &	97.35 &	91.20 &	31.49 &	30.49 &	95.94\\
& bT & F1 & 75.02 &	99.56 &	61.06 &	98.46 &	95.85 &	89.79 &	31.00 &	30.02 &	94.46\\

\bottomrule
 \end{tabular}

%\end{subtable}

\caption{\small Benchmark results for the Morfeusz tagset performed on two datasets: NKJP-\textit{byType} (bT) and NKJP-\textit{byName} (bN); AA -- Aligned Accuracy; F1 -- F1 score. Embeddings used in the models are: \textsl{R} -- xlm-RoBERTa-base, \textsl{fT} -- fastText, \textsl{P} -- Polbert-base, \textsl{pl} -- pl-core-news-lg, \textsl{H} -- HerBERT. }
\label{tab:5FullMorfeusz}
\end{table*}

\begin{table*}[p]
\centering
\small
% \rotatebox[origin=bl]{90}{%
% \begin{subtable}{\textheight}%
% \centering

% \begin{longtable}
% \small
 \begin{tabular}{lc|c|c|c|c|c|c|c|c|c|c}
%   \centering
%  \small
{Model / Task} & \rot{\textbf{Dataset}} &\rot{\textbf{Scores}} &\rot{\textbf{Average}} & \rot{\textbf{Tokens}}  & \rot{\textbf{Sentences}} & \rot{\textbf{Words}}  & \rot{\textbf{UPOS}} & \rot{\textbf{XPOS}} & \rot{\textbf{UFeats}} & \rot{\textbf{AllTags}} & \rot{\textbf{Lemmas}} \\
\midrule
\multirow{1}{*}{\begin{tabular}[c]{@{}l@{}}\textsl{combo} \\ + \textsl{H}\end{tabular}} 
 & bN  & AA & 97.18 & -     & -     & -     & 98.63 & 96.45 & 96.77 & 95.60 & 98.42 \\
 & bN  & F1 & 96.59 & 99.07 & 93.57 & 99.01 & 97.65 & 95.50 & 95.81 & 94.66 & 97.45 \\
 & bT  & AA & 97.17 & -     & -     & -     & 98.66 & 96.46 & 96.70 & 95.57 & 98.48 \\
 & bT  & F1 & 96.58 & 99.15 & 93.08 & 99.07 & 97.75 & 95.56 & 95.80 & 94.68 & 97.57 \\
 & PDB & AA & 93.62 & -     & -     & -     & 99.07 & 96.65 & 96.96 & 96.00 & 98.13 \\
 & PDB & F1 & 93.37 & 99.40 & 96.22 & 98.22 & 97.31 & 94.92 & 95.23 & 94.29 & 96.38 \\
\midrule
\multirow{1}{*}{\begin{tabular}[c]{@{}l@{}}\textsl{stanza} \\ + \textsl{fT}\end{tabular}} 
 & bN  & AA & 94.66 & -     & -     & -     & 97.67 & 93.11 & 93.13 & 91.73 & 97.68 \\
 & bN  & F1 & 95.20 & 99.70 & 92.03 & 99.40 & 97.08 & 92.55 & 92.56 & 91.17 & 97.09 \\
 & bT  & AA & 94.82 & -     & -     & -     & 97.78 & 93.42 & 93.41 & 92.05 & 97.45 \\
 & bT  & F1 & 95.46 & 99.76 & 92.89 & 99.47 & 97.26 & 92.93 & 92.91 & 91.56 & 96.93 \\
 & PDB & AA & 90.60 & -     & -     & -     & 98.21 & 93.71 & 93.76 & 92.69 & 96.32 \\
 & PDB & F1 & 92.10 & 99.86 & 96.83 & 99.42 & 97.64 & 93.17 & 93.22 & 92.15 & 95.77 \\
\midrule
\multirow{1}{*}{\begin{tabular}[c]{@{}l@{}}\textsl{combo} \\ + \textsl{fT}\end{tabular}} 
& bN  & AA & 95.93 & -     & -     & -     & 98.20 & 94.79 & 95.01 & 93.59 & 98.04 \\
 & bN  & F1 & 95.82 & 99.07 & 93.57 & 99.01 & 97.22 & 93.86 & 94.07 & 92.67 & 97.07 \\
 & bT  & AA & 96.00 & -     & -     & -     & 98.28 & 94.88 & 95.05 & 93.69 & 98.07 \\
 & bT  & F1 & 95.85 & 99.13 & 93.08 & 99.07 & 97.37 & 94.00 & 94.17 & 92.83 & 97.16 \\
 & PDB & AA & 89.77 & -     & -     & -     & 97.07 & 93.70 & 93.88 & 92.05 & 97.11 \\
 & PDB & F1 & 90.46 & 99.35 & 96.22 & 98.22 & 95.34 & 92.03 & 92.21 & 90.41 & 95.37 \\
\midrule
\multirow{1}{*}{\begin{tabular}[c]{@{}l@{}}\textsl{trankit} \\ + \textsl{R}\end{tabular}}
& bN  & AA & 92.68 & -     & -     & -     & 97.31 & 91.56 & 91.72 & 89.51 & 93.30 \\
 & bN  & F1 & 92.57 & 98.49 & 90.24 & 97.95 & 95.32 & 89.68 & 89.84 & 87.68 & 91.39 \\
 & bT  & AA & 92.58 & -     & -     & -     & 97.32 & 91.43 & 91.62 & 89.40 & 93.15 \\
 & bT  & F1 & 92.12 & 98.24 & 88.58 & 97.73 & 95.11 & 89.36 & 89.55 & 87.37 & 91.04 \\
 & PDB & AA & 92.51 & -     & -     & -     & 99.18 & 96.28 & 96.44 & 95.68 & 89.08 \\
 & PDB & F1 & 94.03 & 99.90 & 98.51 & 99.89 & 99.07 & 96.18 & 96.34 & 95.57 & 88.98 \\
\midrule
\multirow{1}{*}{\begin{tabular}[c]{@{}l@{}}\textsl{udpipe} \\+ \textsl{fT} \end{tabular}} 
 & bN  & AA & 93.20 & -     & -     & -     & 97.61 & 90.91 & 91.27 & 90.29 & 95.94 \\
 & bN  & F1 & 94.39 & 99.75 & 90.82 & 99.74 & 97.36 & 90.68 & 91.03 & 90.06 & 95.70 \\
 & bT  & AA & 93.17 & -     & -     & -     & 97.59 & 90.88 & 91.24 & 90.20 & 95.94 \\
 & bT  & F1 & 94.35 & 99.77 & 90.59 & 99.76 & 97.35 & 90.65 & 91.02 & 89.98 & 95.70 \\
 & PDB & AA & 85.14 & -     & -     & -     & 97.43 & 88.71 & 89.21 & 88.16 & 94.44 \\
 & PDB & F1 & 88.16 & 99.86 & 95.90 & 99.84 & 97.28 & 88.57 & 89.07 & 88.02 & 94.29 \\
\midrule
\multirow{1}{*}{\begin{tabular}[c]{@{}l@{}}\textsl{spacy} \\ + \textsl{P}\end{tabular}} 
& bN  & AA & 96.82 & -     & -     & -     & 98.63 & 96.37 & 96.50 & 95.72 & 96.87 \\
 & bN  & F1 & 92.15 & 99.56 & 62.64 & 98.45 & 97.10 & 94.88 & 95.00 & 94.23 & 95.37 \\
 & bT  & AA & 96.83 & -     & -     & -     & 98.67 & 96.30 & 96.48 & 95.68 & 97.02 \\
 & bT  & F1 & 91.97 & 99.54 & 61.06 & 98.46 & 97.15 & 94.82 & 94.99 & 94.20 & 95.52 \\
 & PDB & AA & 87.45 & -     & -     & -     & 99.02 & 95.77 & 95.95 & 95.27 & 95.19 \\
 & PDB & F1 & 87.98 & 99.65 & 71.46 & 98.51 & 97.54 & 94.35 & 94.52 & 93.85 & 93.77 \\
\midrule
\multirow{1}{*}{\begin{tabular}[c]{@{}l@{}}\textsl{spacy} \\ + \textsl{pl}\end{tabular}} 
 & bN  & AA & 94.27 & -     & -     & -     & 97.77 & 92.82 & 93.12 & 91.58 & 96.05 \\
 & bN  & F1 & 90.59 & 99.56 & 62.64 & 98.45 & 96.25 & 91.38 & 91.68 & 90.16 & 94.56 \\
 & bT  & AA & 94.24 & -     & -     & -     & 97.84 & 92.75 & 93.01 & 91.50 & 96.10 \\
 & bT  & F1 & 90.38 & 99.54 & 61.06 & 98.46 & 96.34 & 91.32 & 91.57 & 90.09 & 94.62 \\
 & PDB & AA & 82.58 & -     & -     & -     & 97.95 & 91.50 & 91.79 & 90.05 & 93.07 \\
 & PDB & F1 & 84.21 & 99.65 & 71.46 & 98.51 & 96.49 & 90.14 & 90.42 & 88.71 & 91.69 \\
\midrule
\multirow{1}{*}{\begin{tabular}[c]{@{}l@{}}\textsl{spacy} \\ + \textsl{fT}\end{tabular}} 
 & bN  & AA & 93.47 & -     & -     & -     & 97.34 & 91.83 & 92.17 & 90.48 & 95.56 \\
 & bN  & F1 & 90.10 & 99.56 & 62.64 & 98.45 & 95.83 & 90.40 & 90.74 & 89.07 & 94.07 \\
 & bT  & AA & 93.49 & -     & -     & -     & 97.44 & 91.77 & 92.03 & 90.42 & 95.79 \\
 & bT  & F1 & 89.91 & 99.54 & 61.06 & 98.46 & 95.93 & 90.35 & 90.62 & 89.03 & 94.32 \\
 & PDB & AA & 81.07 & -     & -     & -     & 97.06 & 89.91 & 90.13 & 88.31 & 93.10 \\
 & PDB & F1 & 83.03 & 99.65 & 71.46 & 98.51 & 95.62 & 88.57 & 88.79 & 87.00 & 91.72 \\

\bottomrule
 \end{tabular}

% \end{subtable}%\qquad
% }% end of \rotatebox

\caption{\small Benchmark results for the UD tagset performed on three datasets: NKJP-\textit{byType} (bT), NKJP-\textit{byName} (bN), and PDB-UD (PDB) for segmentation, tagging and lemmatization tasks; AA -- Aligned Accuracy; F1 -- F1 score. Embeddings used in the models are: \textsl{R} -- xlm-RoBERTa-base, \textsl{fT} -- fastText, \textsl{P} -- Polbert-base, \textsl{pl} -- pl-core-news-lg, \textsl{H} -- HerBERT-base. }
\label{tab:6UDonTagging}
% \end{longtable}
\end{table*}

\begin{table*}[p]
\centering
\small
% \rotatebox[origin=bl]{90}{%
 %\begin{subtable}{\textheight}%
% \centering

% \begin{longtable}
% \small
 \begin{tabular}{lc|c|c|c|c|c|c|c}
%   \centering
%  \small
{Model / Task} & \rot{\textbf{Dataset}} &\rot{\textbf{Scores}} &\rot{\textbf{Average}} & \rot{\textbf{UAS}}  & \rot{\textbf{LAS}} & \rot{\textbf{CLAS}}  & \rot{\textbf{MLAS}} & \rot{\textbf{BLEX}} \\
\midrule
\multirow{1}{*}{\begin{tabular}[c]{@{}l@{}}\textsl{combo} \\ + \textsl{H}\end{tabular}} 
& bN  & AA & -     & -     & -     & -     & -     \\
 & bN  & F1 & -     & -     & -     & -     & -     \\
 & bT  & AA & -     & -     & -     & -     & -     \\
 & bT  & F1 & -     & -     & -     & -     & -     \\
 & PDB & AA & 93.62 & 92.97 & 91.61 & 90.47 & 86.15 & 88.18 \\
 & PDB & F1 &93.37 & 91.31 & 89.98 & 89.03 & 84.77 & 86.77 \\
\midrule
\multirow{1}{*}{\begin{tabular}[c]{@{}l@{}}\textsl{stanza} \\ + \textsl{fT}\end{tabular}} 
& bN  & AA & -     & -     & -     & -     & -     \\
 & bN  & F1 & -     & -     & -     & -     & -     \\
 & bT  & AA & -     & -     & -     & -     & -     \\
 & bT  & F1 & -     & -     & -     & -     & -     \\
 & PDB & AA & 90.60 & 91.62 & 89.34 & 87.25 & 80.22 & 82.87 \\
 & PDB & F1 & 92.10 & 91.09 & 88.83 & 86.90 & 79.90 & 82.53 \\
\midrule
\multirow{1}{*}{\begin{tabular}[c]{@{}l@{}}\textsl{combo} \\ + \textsl{fT}\end{tabular}} 
& bN  & AA & -     & -     & -     & -     & -     \\
 & bN  & F1 & -     & -     & -     & -     & -     \\
 & bT  & AA & -     & -     & -     & -     & -     \\
 & bT  & F1 & -     & -     & -     & -     & -     \\
 & PDB & AA &  89.77 & 90.10 & 87.76 & 85.49 & 77.88 & 82.65 \\
 & PDB & F1 & 90.46 & 88.49 & 86.19 & 84.14 & 76.64 & 81.34 \\
\midrule
\multirow{1}{*}{\begin{tabular}[c]{@{}l@{}}\textsl{trankit} \\ + \textsl{R}\end{tabular}}
& bN  & AA & -     & -     & -     & -     & -     \\
 & bN  & F1 & -     & -     & -     & -     & -     \\
 & bT  & AA & -     & -     & -     & -     & -     \\
 & bT  & F1 & -     & -     & -     & -     & -     \\
 & PDB & AA & 92.51 & 95.89 & 94.34 & 93.10 & 87.88 & 77.26 \\
 & PDB & F1 & 94.03 & 95.79 & 94.24 & 93.00 & 87.79 & 77.18 \\
\midrule
\multirow{1}{*}{\begin{tabular}[c]{@{}l@{}}\textsl{udpipe} \\ + \textsl{fT}\end{tabular}} 
& bN  & AA & -     & -     & -     & -     & -     \\
 & bN  & F1 & -     & -     & -     & -     & -     \\
 & bT  & AA & -     & -     & -     & -     & -     \\
 & bT  & F1 & -     & -     & -     & -     & -     \\
 & PDB & AA & 85.14 & 86.82 & 83.14 & 79.52 & 69.52 & 74.48 \\
 & PDB & F1 & 88.16 & 86.68 & 83.01 & 79.53 & 69.53 & 74.49 \\
\midrule
\multirow{1}{*}{\begin{tabular}[c]{@{}l@{}}\textsl{spacy} \\+ \textsl{P} \end{tabular}} 
 & bN  & AA & -     & -     & -     & -     & -     \\
 & bN  & F1 & -     & -     & -     & -     & -     \\
 & bT  & AA & -     & -     & -     & -     & -     \\
 & bT  & F1 & -     & -     & -     & -     & -     \\
 & PDB & AA & 87.45 & 89.41 & 81.54 & 77.23 & 72.67 & 72.41 \\
 & PDB & F1 & 87.98 & 88.08 & 80.33 & 80.50 & 75.75 & 75.48 \\
\midrule
\multirow{1}{*}{\begin{tabular}[c]{@{}l@{}}\textsl{spacy} \\ + \textsl{pl}\end{tabular}} 
  & bN  & AA & -     & -     & -     & -     & -     \\
 & bN  & F1 & -     & -     & -     & -     & -     \\
 & bT  & AA & -     & -     & -     & -     & -     \\
 & bT  & F1 & -     & -     & -     & -     & -     \\
 & PDB & AA & 82.58 & 83.39 & 74.71 & 72.66 & 64.21 & 66.50 \\
 & PDB & F1 & 84.21 & 82.15 & 73.60 & 75.71 & 66.90 & 69.29 \\
\midrule
\multirow{1}{*}{\begin{tabular}[c]{@{}l@{}}\textsl{spacy} \\ + \textsl{fT}\end{tabular}} 
 & bN  & AA & -     & -     & -     & -     & -     \\
 & bN  & F1 & -     & -     & -     & -     & -     \\
 & bT  & AA & -     & -     & -     & -     & -     \\
 & bT  & F1 & -     & -     & -     & -     & -     \\
 & PDB & AA & 81.07 & 82.13 & 73.33 & 70.76 & 61.02 & 64.93 \\
 & PDB & F1 & 83.03 & 80.91 & 72.24 & 73.72 & 63.57 & 67.64 \\

\bottomrule
 \end{tabular}

 %\end{subtable}%\qquad
% }% end of \rotatebox

\caption{\small Benchmark results for the UD tagset performed on three datasets: NKJP-\textit{byType} (bT), NKJP-\textit{byName} (bN), and PDB-UD (PDB) for the dependency parsing task; AA -- Aligned Accuracy; F1 -- F1 score. Embeddings used in the models are: \textsl{R} -- xlm-RoBERTa-base, \textsl{fT} -- fastText, \textsl{P} -- Polbert-base, \textsl{pl} -- pl-core-news-lg, \textsl{H} -- HerBERT. }
\label{tab:7UDonparsing}
% \end{longtable}
\end{table*}

\section{Acknowledgements}
This work was supported by the European Regional Development Fund as a part of 2014–2020 Smart Growth Operational Programme, CLARIN — Common Language Resources and Technology Infrastructure (project no. POIR.04.02.00-00C002/19) and DARIAH-PL — Digital Research Infrastructure for the Arts and Humanities (project no. POIR.04.02.00-00-D006/20-0), and as part of the investment CLARIN ERIC: Common Language Resources and Technology Infrastructure (period: 2024-2026) funded by the Polish Ministry of Science and Higher Education (agreement no. 2024/WK/01). We gratefully acknowledge Poland’s high-performance computing infrastructure PLGrid (HPC Centers: ACK Cyfronet AGH) for providing computer facilities and support within computational grant no. PLG/2022/015872.

\nocite{bojanowski-etal-2017-enriching,conneau-etal-2020-unsupervised,mroczkowski-etal-2021-herbert}
\section{Bibliographical References}
\label{sec:reference}

\bibliographystyle{lrec-coling2024-natbib}
\bibliography{custom}

\begin{thebibliography}{47}
\expandafter\ifx\csname natexlab\endcsname\relax\def\natexlab#1{#1}\fi

\bibitem[{Bojanowski et~al.(2017)Bojanowski, Grave, Joulin, and Mikolov}]{bojanowski-etal-2017-enriching}
Piotr Bojanowski, Edouard Grave, Armand Joulin, and Tomas Mikolov. 2017.
\newblock \href {https://doi.org/10.1162/tacl_a_00051} {Enriching word vectors with subword information}.
\newblock \emph{Transactions of the Association for Computational Linguistics}, 5:135--146.

\bibitem[{Brown et~al.(2020)Brown, Mann, Ryder, Subbiah, Kaplan, Dhariwal, Neelakantan, Shyam, Sastry, Askell, Agarwal, Herbert-Voss, Krueger, Henighan, Child, Ramesh, Ziegler, Wu, Winter, Hesse, Chen, Sigler, Litwin, Gray, Chess, Clark, Berner, McCandlish, Radford, Sutskever, and Amodei}]{brown:2020}
Tom Brown, Benjamin Mann, Nick Ryder, Melanie Subbiah, Jared~D Kaplan, Prafulla Dhariwal, Arvind Neelakantan, Pranav Shyam, Girish Sastry, Amanda Askell, Sandhini Agarwal, Ariel Herbert-Voss, Gretchen Krueger, Tom Henighan, Rewon Child, Aditya Ramesh, Daniel Ziegler, Jeffrey Wu, Clemens Winter, Chris Hesse, Mark Chen, Eric Sigler, Mateusz Litwin, Scott Gray, Benjamin Chess, Jack Clark, Christopher Berner, Sam McCandlish, Alec Radford, Ilya Sutskever, and Dario Amodei. 2020.
\newblock \href {https://proceedings.neurips.cc/paper_files/paper/2020/file/1457c0d6bfcb4967418bfb8ac142f64a-Paper.pdf} {{Language Models are Few-Shot Learners}}.
\newblock In \emph{Advances in Neural Information Processing Systems}, volume~33, pages 1877--1901. Curran Associates, Inc.

\bibitem[{Buchholz and Marsi(2006)}]{buchholz-marsi-2006-conll}
Sabine Buchholz and Erwin Marsi. 2006.
\newblock \href {https://aclanthology.org/W06-2920} {{C}o{NLL}-{X} shared task on multilingual dependency parsing}.
\newblock In \emph{Proceedings of the Tenth Conference on Computational Natural Language Learning ({C}o{NLL}-X)}, pages 149--164, New York City. Association for Computational Linguistics.

\bibitem[{Chen et~al.(2017)Chen, Zhao, Yang, and Liu}]{Chen_Zhao_Yang_Liu_2017}
Kehai Chen, Tiejun Zhao, Muyun Yang, and Lemao Liu. 2017.
\newblock \href {https://doi.org/10.1609/aaai.v31i1.10978} {Translation prediction with source dependency-based context representation}.
\newblock \emph{Proceedings of the AAAI Conference on Artificial Intelligence}, 31(1).

\bibitem[{Conneau et~al.(2020)Conneau, Khandelwal, Goyal, Chaudhary, Wenzek, Guzm{\'a}n, Grave, Ott, Zettlemoyer, and Stoyanov}]{conneau-etal-2020-unsupervised}
Alexis Conneau, Kartikay Khandelwal, Naman Goyal, Vishrav Chaudhary, Guillaume Wenzek, Francisco Guzm{\'a}n, Edouard Grave, Myle Ott, Luke Zettlemoyer, and Veselin Stoyanov. 2020.
\newblock \href {https://doi.org/10.18653/v1/2020.acl-main.747} {Unsupervised cross-lingual representation learning at scale}.
\newblock In \emph{Proceedings of the 58th Annual Meeting of the Association for Computational Linguistics}, pages 8440--8451, Online. Association for Computational Linguistics.

\bibitem[{Crouch et~al.(2011)Crouch, Dalrymple, Kaplan, King, Maxwell, and Newman}]{xle}
Dick Crouch, Mary Dalrymple, Ronald~M. Kaplan, Tracy~Holloway King, John Maxwell, and Paula Newman. 2011.
\newblock \href {https://ling.sprachwiss.uni-konstanz.de/pages/xle/doc/xle_toc.html} {{XLE Documentation}}.
\newblock Palo Alto Research Center.

\bibitem[{de~Marneffe et~al.(2021)de~Marneffe, Manning, Nivre, and Zeman}]{de-marneffe-etal-2021-universal}
Marie-Catherine de~Marneffe, Christopher~D. Manning, Joakim Nivre, and Daniel Zeman. 2021.
\newblock \href {https://doi.org/10.1162/coli_a_00402} {{U}niversal {D}ependencies}.
\newblock \emph{Computational Linguistics}, 47(2):255--308.

\bibitem[{Gehrmann et~al.(2021)Gehrmann, Adewumi, Aggarwal, Ammanamanchi, Aremu, Bosselut, Chandu, Clinciu, Das, Dhole, Du, Durmus, Du{\v{s}}ek, Emezue, Gangal, Garbacea, Hashimoto, Hou, Jernite, Jhamtani, Ji, Jolly, Kale, Kumar, Ladhak, Madaan, Maddela, Mahajan, Mahamood, Majumder, Martins, McMillan-Major, Mille, van Miltenburg, Nadeem, Narayan, Nikolaev, Niyongabo~Rubungo, Osei, Parikh, Perez-Beltrachini, Rao, Raunak, Rodriguez, Santhanam, Sedoc, Sellam, Shaikh, Shimorina, Sobrevilla~Cabezudo, Strobelt, Subramani, Xu, Yang, Yerukola, and Zhou}]{gehrmann-etal-2021-gem}
Sebastian Gehrmann, Tosin Adewumi, Karmanya Aggarwal, Pawan~Sasanka Ammanamanchi, Anuoluwapo Aremu, Antoine Bosselut, Khyathi~Raghavi Chandu, Miruna-Adriana Clinciu, Dipanjan Das, Kaustubh Dhole, Wanyu Du, Esin Durmus, Ond{\v{r}}ej Du{\v{s}}ek, Chris~Chinenye Emezue, Varun Gangal, Cristina Garbacea, Tatsunori Hashimoto, Yufang Hou, Yacine Jernite, Harsh Jhamtani, Yangfeng Ji, Shailza Jolly, Mihir Kale, Dhruv Kumar, Faisal Ladhak, Aman Madaan, Mounica Maddela, Khyati Mahajan, Saad Mahamood, Bodhisattwa~Prasad Majumder, Pedro~Henrique Martins, Angelina McMillan-Major, Simon Mille, Emiel van Miltenburg, Moin Nadeem, Shashi Narayan, Vitaly Nikolaev, Andre Niyongabo~Rubungo, Salomey Osei, Ankur Parikh, Laura Perez-Beltrachini, Niranjan~Ramesh Rao, Vikas Raunak, Juan~Diego Rodriguez, Sashank Santhanam, Jo{\~a}o Sedoc, Thibault Sellam, Samira Shaikh, Anastasia Shimorina, Marco~Antonio Sobrevilla~Cabezudo, Hendrik Strobelt, Nishant Subramani, Wei Xu, Diyi Yang, Akhila Yerukola, and Jiawei Zhou. 2021.
\newblock \href {https://doi.org/10.18653/v1/2021.gem-1.10} {The {GEM} benchmark: Natural language generation, its evaluation and metrics}.
\newblock In \emph{Proceedings of the 1st Workshop on Natural Language Generation, Evaluation, and Metrics (GEM 2021)}, pages 96--120, Online. Association for Computational Linguistics.

\bibitem[{Graves and Schmidhuber(2005)}]{GRAVES2005602}
Alex Graves and Jürgen Schmidhuber. 2005.
\newblock \href {https://doi.org/https://doi.org/10.1016/j.neunet.2005.06.042} {Framewise phoneme classification with bidirectional lstm and other neural network architectures}.
\newblock \emph{Neural Networks}, 18(5):602--610.
\newblock IJCNN 2005.

\bibitem[{Guo et~al.(2019)Guo, Zhang, and Lu}]{guo-etal-2019-attention}
Zhijiang Guo, Yan Zhang, and Wei Lu. 2019.
\newblock \href {https://doi.org/10.18653/v1/P19-1024} {Attention guided graph convolutional networks for relation extraction}.
\newblock In \emph{Proceedings of the 57th Annual Meeting of the Association for Computational Linguistics}, pages 241--251, Florence, Italy. Association for Computational Linguistics.

\bibitem[{Hu et~al.(2020)Hu, Ruder, Siddhant, Neubig, Firat, and Johnson}]{pmlr-v119-hu20b}
Junjie Hu, Sebastian Ruder, Aditya Siddhant, Graham Neubig, Orhan Firat, and Melvin Johnson. 2020.
\newblock \href {https://proceedings.mlr.press/v119/hu20b.html} {{XTREME}: A massively multilingual multi-task benchmark for evaluating cross-lingual generalisation}.
\newblock In \emph{Proceedings of the 37th International Conference on Machine Learning}, volume 119 of \emph{Proceedings of Machine Learning Research}, pages 4411--4421. PMLR.

\bibitem[{Kasai et~al.(2019)Kasai, Friedman, Frank, Radev, and Rambow}]{kasai-etal-2019-syntax}
Jungo Kasai, Dan Friedman, Robert Frank, Dragomir Radev, and Owen Rambow. 2019.
\newblock \href {https://doi.org/10.18653/v1/N19-1075} {Syntax-aware neural semantic role labeling with supertags}.
\newblock In \emph{Proceedings of the 2019 Conference of the North {A}merican Chapter of the Association for Computational Linguistics: Human Language Technologies, Volume 1 (Long and Short Papers)}, pages 701--709, Minneapolis, Minnesota. Association for Computational Linguistics.

\bibitem[{Khashabi et~al.(2018)Khashabi, Khot, Sabharwal, and Roth}]{Khashabi_Khot_Sabharwal_Roth_2018}
Daniel Khashabi, Tushar Khot, Ashish Sabharwal, and Dan Roth. 2018.
\newblock \href {https://doi.org/10.1609/aaai.v32i1.11574} {Question answering as global reasoning over semantic abstractions}.
\newblock \emph{Proceedings of the AAAI Conference on Artificial Intelligence}, 32(1).

\bibitem[{Kiela et~al.(2021)Kiela, Bartolo, Nie, Kaushik, Geiger, Wu, Vidgen, Prasad, Singh, Ringshia, Ma, Thrush, Riedel, Waseem, Stenetorp, Jia, Bansal, Potts, and Williams}]{kiela-etal-2021-dynabench}
Douwe Kiela, Max Bartolo, Yixin Nie, Divyansh Kaushik, Atticus Geiger, Zhengxuan Wu, Bertie Vidgen, Grusha Prasad, Amanpreet Singh, Pratik Ringshia, Zhiyi Ma, Tristan Thrush, Sebastian Riedel, Zeerak Waseem, Pontus Stenetorp, Robin Jia, Mohit Bansal, Christopher Potts, and Adina Williams. 2021.
\newblock \href {https://doi.org/10.18653/v1/2021.naacl-main.324} {Dynabench: Rethinking benchmarking in {NLP}}.
\newblock In \emph{Proceedings of the 2021 Conference of the North American Chapter of the Association for Computational Linguistics: Human Language Technologies}, pages 4110--4124, Online. Association for Computational Linguistics.

\bibitem[{Kieraś and Woliński(2017)}]{kie:wol:17:morf}
Witold Kieraś and Marcin Woliński. 2017.
\newblock Morfeusz 2 – analizator i~generator fleksyjny dla języka polskiego.
\newblock \emph{Język Polski}, XCVII(1):75--83.

\bibitem[{Klimaszewski and Wr{\'o}blewska(2021)}]{klimaszewski-wroblewska-2021-combo-state}
Mateusz Klimaszewski and Alina Wr{\'o}blewska. 2021.
\newblock \href {https://doi.org/10.18653/v1/2021.emnlp-demo.7} {{COMBO}: State-of-the-art morphosyntactic analysis}.
\newblock In \emph{Proceedings of the 2021 Conference on Empirical Methods in Natural Language Processing: System Demonstrations}, pages 50--62, Online and Punta Cana, Dominican Republic. Association for Computational Linguistics.

\bibitem[{McDonald et~al.(2005)McDonald, Crammer, and Pereira}]{mcdonald-etal-2005-online}
Ryan McDonald, Koby Crammer, and Fernando Pereira. 2005.
\newblock \href {https://doi.org/10.3115/1219840.1219852} {Online large-margin training of dependency parsers}.
\newblock In \emph{Proceedings of the 43rd Annual Meeting of the Association for Computational Linguistics ({ACL}{'}05)}, pages 91--98, Ann Arbor, Michigan. Association for Computational Linguistics.

\bibitem[{Montani and Honnibal(2022)}]{spacy3}
Ines Montani and Matthew Honnibal. 2022.
\newblock \href {https://spacy.io} {{spaCy: Industrial-Strength Natural Language Processing in Python}}.
\newblock Version 3.4.1.

\bibitem[{Mroczkowski et~al.(2021)Mroczkowski, Rybak, Wr{\'o}blewska, and Gawlik}]{mroczkowski-etal-2021-herbert}
Robert Mroczkowski, Piotr Rybak, Alina Wr{\'o}blewska, and Ireneusz Gawlik. 2021.
\newblock \href {https://aclanthology.org/2021.bsnlp-1.1} {{H}er{BERT}: Efficiently pretrained transformer-based language model for {P}olish}.
\newblock In \emph{Proceedings of the 8th Workshop on Balto-Slavic Natural Language Processing}, pages 1--10, Kiyv, Ukraine. Association for Computational Linguistics.

\bibitem[{Nguyen et~al.(2021{\natexlab{a}})Nguyen, Lai, Pouran Ben~Veyseh, and Nguyen}]{nguyen-etal-2021-trankit}
Minh~Van Nguyen, Viet~Dac Lai, Amir Pouran Ben~Veyseh, and Thien~Huu Nguyen. 2021{\natexlab{a}}.
\newblock \href {https://doi.org/10.18653/v1/2021.eacl-demos.10} {Trankit: A light-weight transformer-based toolkit for multilingual natural language processing}.
\newblock In \emph{Proceedings of the 16th Conference of the European Chapter of the Association for Computational Linguistics: System Demonstrations}, pages 80--90, Online. Association for Computational Linguistics.

\bibitem[{Nguyen et~al.(2021{\natexlab{b}})Nguyen, Lai, Veyseh, and Nguyen}]{nguyen2021trankit}
Minh~Van Nguyen, Viet~Dac Lai, Amir Pouran~Ben Veyseh, and Thien~Huu Nguyen. 2021{\natexlab{b}}.
\newblock Trankit: A light-weight transformer-based toolkit for multilingual natural language processing.
\newblock In \emph{Proceedings of the 16th Conference of the European Chapter of the Association for Computational Linguistics: System Demonstrations}.

\bibitem[{Nivre(2009)}]{nivre-2009-non}
Joakim Nivre. 2009.
\newblock \href {https://aclanthology.org/P09-1040} {Non-projective dependency parsing in expected linear time}.
\newblock In \emph{Proceedings of the Joint Conference of the 47th Annual Meeting of the {ACL} and the 4th International Joint Conference on Natural Language Processing of the {AFNLP}}, pages 351--359, Suntec, Singapore. Association for Computational Linguistics.

\bibitem[{Ouyang et~al.(2022)Ouyang, Wu, Jiang, Almeida, Wainwright, Mishkin, Zhang, Agarwal, Slama, Ray, Schulman, Hilton, Kelton, Miller, Simens, Askell, Welinder, Christiano, Leike, and Lowe}]{ouyang2022training}
Long Ouyang, Jeff Wu, Xu~Jiang, Diogo Almeida, Carroll~L. Wainwright, Pamela Mishkin, Chong Zhang, Sandhini Agarwal, Katarina Slama, Alex Ray, John Schulman, Jacob Hilton, Fraser Kelton, Luke Miller, Maddie Simens, Amanda Askell, Peter Welinder, Paul Christiano, Jan Leike, and Ryan Lowe. 2022.
\newblock \href {http://arxiv.org/abs/2203.02155} {Training language models to follow instructions with human feedback}.

\bibitem[{Pavao et~al.(2022)Pavao, Guyon, Letournel, Baró, Escalante, Escalera, Thomas, and Xu}]{codalab_competitions}
Adrien Pavao, Isabelle Guyon, Anne-Catherine Letournel, Xavier Baró, Hugo Escalante, Sergio Escalera, Tyler Thomas, and Zhen Xu. 2022.
\newblock \href {https://hal.inria.fr/hal-03629462v1} {Codalab competitions: An open source platform to organize scientific challenges}.
\newblock Technical report, Université Paris-Saclay.

\bibitem[{Pfeiffer et~al.(2020{\natexlab{a}})Pfeiffer, R{\"u}ckl{\'e}, Poth, Kamath, Vuli{\'c}, Ruder, Cho, and Gurevych}]{pfeiffer-etal-2020-adapterhub}
Jonas Pfeiffer, Andreas R{\"u}ckl{\'e}, Clifton Poth, Aishwarya Kamath, Ivan Vuli{\'c}, Sebastian Ruder, Kyunghyun Cho, and Iryna Gurevych. 2020{\natexlab{a}}.
\newblock \href {https://doi.org/10.18653/v1/2020.emnlp-demos.7} {{A}dapter{H}ub: A framework for adapting transformers}.
\newblock In \emph{Proceedings of the 2020 Conference on Empirical Methods in Natural Language Processing: System Demonstrations}, pages 46--54, Online. Association for Computational Linguistics.

\bibitem[{Pfeiffer et~al.(2020{\natexlab{b}})Pfeiffer, Vuli{\'c}, Gurevych, and Ruder}]{pfeiffer-etal-2020-mad}
Jonas Pfeiffer, Ivan Vuli{\'c}, Iryna Gurevych, and Sebastian Ruder. 2020{\natexlab{b}}.
\newblock \href {https://doi.org/10.18653/v1/2020.emnlp-main.617} {{MAD-X}: {A}n {A}dapter-{B}ased {F}ramework for {M}ulti-{T}ask {C}ross-{L}ingual {T}ransfer}.
\newblock In \emph{Proceedings of the 2020 Conference on Empirical Methods in Natural Language Processing (EMNLP)}, pages 7654--7673, Online. Association for Computational Linguistics.

\bibitem[{Przepiórkowski et~al.(2012)Przepiórkowski, Bańko, Górski, and Lewandowska-Tomaszczyk}]{prz:etal:11:ed}
Adam Przepiórkowski, Mirosław Bańko, Rafał~L. Górski, and Barbara Lewandowska-Tomaszczyk, editors. 2012.
\newblock \emph{{N}arodowy {K}orpus {J}ęzyka {P}olskiego}.
\newblock Wydawnictwo Naukowe PWN, Warsaw.

\bibitem[{Przyby{\l}a(2022)}]{lambo}
Piotr Przyby{\l}a. 2022.
\newblock \href {https://gitlab.clarin-pl.eu/syntactic-tools/lambo} {{LAMBO: Layered Approach to Multi-level BOundary identification}}.

\bibitem[{Qi et~al.(2020)Qi, Zhang, Zhang, Bolton, and Manning}]{qi-etal-2020-stanza}
Peng Qi, Yuhao Zhang, Yuhui Zhang, Jason Bolton, and Christopher~D. Manning. 2020.
\newblock \href {https://doi.org/10.18653/v1/2020.acl-demos.14} {{S}tanza: A python natural language processing toolkit for many human languages}.
\newblock In \emph{Proceedings of the 58th Annual Meeting of the Association for Computational Linguistics: System Demonstrations}, pages 101--108, Online. Association for Computational Linguistics.

\bibitem[{Rybak and Wr{\'o}blewska(2018)}]{rybak-wroblewska-2018-semi}
Piotr Rybak and Alina Wr{\'o}blewska. 2018.
\newblock \href {https://doi.org/10.18653/v1/K18-2004} {Semi-supervised neural system for tagging, parsing and lematization}.
\newblock In \emph{Proceedings of the {C}o{NLL} 2018 Shared Task: Multilingual Parsing from Raw Text to Universal Dependencies}, pages 45--54, Brussels, Belgium. Association for Computational Linguistics.

\bibitem[{Sachan et~al.(2021)Sachan, Zhang, Qi, and Hamilton}]{sachan-etal-2021-syntax}
Devendra Sachan, Yuhao Zhang, Peng Qi, and William~L. Hamilton. 2021.
\newblock \href {https://doi.org/10.18653/v1/2021.eacl-main.228} {Do syntax trees help pre-trained transformers extract information?}
\newblock In \emph{Proceedings of the 16th Conference of the European Chapter of the Association for Computational Linguistics: Main Volume}, pages 2647--2661, Online. Association for Computational Linguistics.

\bibitem[{Seddah et~al.(2013)Seddah, Tsarfaty, K{\"u}bler, Candito, Choi, Farkas, Foster, Goenaga, Gojenola~Galletebeitia, Goldberg, Green, Habash, Kuhlmann, Maier, Nivre, Przepi{\'o}rkowski, Roth, Seeker, Versley, Vincze, Woli{\'n}ski, Wr{\'o}blewska, and Villemonte de~la Clergerie}]{seddah-etal-2013-overview}
Djam{\'e} Seddah, Reut Tsarfaty, Sandra K{\"u}bler, Marie Candito, Jinho~D. Choi, Rich{\'a}rd Farkas, Jennifer Foster, Iakes Goenaga, Koldo Gojenola~Galletebeitia, Yoav Goldberg, Spence Green, Nizar Habash, Marco Kuhlmann, Wolfgang Maier, Joakim Nivre, Adam Przepi{\'o}rkowski, Ryan Roth, Wolfgang Seeker, Yannick Versley, Veronika Vincze, Marcin Woli{\'n}ski, Alina Wr{\'o}blewska, and Eric Villemonte de~la Clergerie. 2013.
\newblock \href {https://aclanthology.org/W13-4917} {Overview of the {SPMRL} 2013 shared task: A cross-framework evaluation of parsing morphologically rich languages}.
\newblock In \emph{Proceedings of the Fourth Workshop on Statistical Parsing of Morphologically-Rich Languages}, pages 146--182, Seattle, Washington, USA. Association for Computational Linguistics.

\bibitem[{Straka et~al.(2016)Straka, Haji{\v{c}}, and Strakov{\'a}}]{straka-etal-2016-udpipe}
Milan Straka, Jan Haji{\v{c}}, and Jana Strakov{\'a}. 2016.
\newblock \href {https://aclanthology.org/L16-1680} {{UDP}ipe: Trainable pipeline for processing {C}o{NLL}-{U} files performing tokenization, morphological analysis, {POS} tagging and parsing}.
\newblock In \emph{Proceedings of the Tenth International Conference on Language Resources and Evaluation ({LREC}'16)}, pages 4290--4297, Portoro{\v{z}}, Slovenia. European Language Resources Association (ELRA).

\bibitem[{Straka and Strakov\'{a}(2017)}]{udpipe:2017}
Milan Straka and Jana Strakov\'{a}. 2017.
\newblock \href {http://www.aclweb.org/anthology/K/K17/K17-3009.pdf} {Tokenizing, pos tagging, lemmatizing and parsing ud 2.0 with udpipe}.
\newblock In \emph{Proceedings of the CoNLL 2017 Shared Task: Multilingual Parsing from Raw Text to Universal Dependencies}, pages 88--99, Vancouver, Canada. Association for Computational Linguistics.

\bibitem[{Sun et~al.(2019)Sun, Zhang, Mensah, Mao, and Liu}]{sun-etal-2019-aspect}
Kai Sun, Richong Zhang, Samuel Mensah, Yongyi Mao, and Xudong Liu. 2019.
\newblock \href {https://doi.org/10.18653/v1/D19-1569} {Aspect-level sentiment analysis via convolution over dependency tree}.
\newblock In \emph{Proceedings of the 2019 Conference on Empirical Methods in Natural Language Processing and the 9th International Joint Conference on Natural Language Processing (EMNLP-IJCNLP)}, pages 5679--5688, Hong Kong, China. Association for Computational Linguistics.

\bibitem[{Szałkiewicz and Przepiórkowski(2012)}]{sza:prz:11}
Łukasz Szałkiewicz and Adam Przepiórkowski. 2012.
\newblock Anotacja morfoskładniowa.
\newblock In Adam Przepiórkowski, Mirosław Bańko, Rafał~L. Górski, and Barbara Lewandowska-Tomaszczyk, editors, \emph{{N}arodowy {K}orpus {J}ęzyka {P}olskiego}, pages 59--96. Wydawnictwo Naukowe PWN, Warsaw.

\bibitem[{Vashishth et~al.(2018)Vashishth, Joshi, Prayaga, Bhattacharyya, and Talukdar}]{vashishth-etal-2018-reside}
Shikhar Vashishth, Rishabh Joshi, Sai~Suman Prayaga, Chiranjib Bhattacharyya, and Partha Talukdar. 2018.
\newblock \href {https://doi.org/10.18653/v1/D18-1157} {{RESIDE}: Improving distantly-supervised neural relation extraction using side information}.
\newblock In \emph{Proceedings of the 2018 Conference on Empirical Methods in Natural Language Processing}, pages 1257--1266, Brussels, Belgium. Association for Computational Linguistics.

\bibitem[{Wang et~al.(2018)Wang, Singh, Michael, Hill, Levy, and Bowman}]{wang-etal-2018-glue}
Alex Wang, Amanpreet Singh, Julian Michael, Felix Hill, Omer Levy, and Samuel Bowman. 2018.
\newblock \href {https://doi.org/10.18653/v1/W18-5446} {{GLUE}: A multi-task benchmark and analysis platform for natural language understanding}.
\newblock In \emph{Proceedings of the 2018 {EMNLP} Workshop {B}lackbox{NLP}: Analyzing and Interpreting Neural Networks for {NLP}}, pages 353--355, Brussels, Belgium. Association for Computational Linguistics.

\bibitem[{Wang et~al.(2019)Wang, Johnson, Wan, Sun, and Wang}]{wang-etal-2019-best}
Yufei Wang, Mark Johnson, Stephen Wan, Yifang Sun, and Wei Wang. 2019.
\newblock \href {https://doi.org/10.18653/v1/P19-1529} {How to best use syntax in semantic role labelling}.
\newblock In \emph{Proceedings of the 57th Annual Meeting of the Association for Computational Linguistics}, pages 5338--5343, Florence, Italy. Association for Computational Linguistics.

\bibitem[{Waszczuk(2012)}]{waszczuk2012harnessing}
Jakub Waszczuk. 2012.
\newblock Harnessing the crf complexity with domain-specific constraints. the case of morphosyntactic tagging of a highly inflected language.
\newblock In \emph{Proceedings of COLING 2012}, pages 2789--2804.

\bibitem[{Waszczuk et~al.(2018)Waszczuk, Kiera{\'s}, and Woli{\'n}ski}]{waszczuk2018morphosyntactic}
Jakub Waszczuk, Witold Kiera{\'s}, and Marcin Woli{\'n}ski. 2018.
\newblock Morphosyntactic disambiguation and segmentation for historical polish with graph-based conditional random fields.
\newblock In \emph{International Conference on Text, Speech, and Dialogue}, pages 188--196. Springer.

\bibitem[{Woliński(2014)}]{wol:14}
Marcin Woliński. 2014.
\newblock \href {http://www.lrec-conf.org/proceedings/lrec2014/index.html} {Morfeusz reloaded}.
\newblock In \emph{Proceedings of the Ninth International {C}onference on {L}anguage {R}esources and {E}valuation, {LREC}~2014}, pages 1106--1111. European Language Resources Association (ELRA).

\bibitem[{Woliński(2019)}]{woli:19:wuw}
Marcin Woliński. 2019.
\newblock \href {https://www.wuw.pl/data/include/cms/Automatyczna_analiza_skladnikowa_Wolinski_Marcin_2019.pdf} {\emph{Automatyczna analiza składnikowa języka polskiego}}.
\newblock Wydawnictwa Uniwersytetu Warszawskiego, Warsaw.

\bibitem[{Zeman et~al.(2018)Zeman, Haji{\v{c}}, Popel, Potthast, Straka, Ginter, Nivre, and Petrov}]{zeman-etal-2018-conll}
Daniel Zeman, Jan Haji{\v{c}}, Martin Popel, Martin Potthast, Milan Straka, Filip Ginter, Joakim Nivre, and Slav Petrov. 2018.
\newblock \href {https://doi.org/10.18653/v1/K18-2001} {{C}o{NLL} 2018 shared task: Multilingual parsing from raw text to {U}niversal {D}ependencies}.
\newblock In \emph{Proceedings of the {C}o{NLL} 2018 Shared Task: Multilingual Parsing from Raw Text to Universal Dependencies}, pages 1--21, Brussels, Belgium. Association for Computational Linguistics.

\bibitem[{Zeman et~al.(2017)Zeman, Popel, Straka, Haji{\v{c}}, Nivre, Ginter, Luotolahti, Pyysalo, Petrov, Potthast, Tyers, Badmaeva, Gokirmak, Nedoluzhko, Cinkov{\'a}, Haji{\v{c}}~jr., Hlav{\'a}{\v{c}}ov{\'a}, Kettnerov{\'a}, Ure{\v{s}}ov{\'a}, Kanerva, Ojala, Missil{\"a}, Manning, Schuster, Reddy, Taji, Habash, Leung, de~Marneffe, Sanguinetti, Simi, Kanayama, de~Paiva, Droganova, Mart{\'\i}nez~Alonso, {\c{C}}{\"o}ltekin, Sulubacak, Uszkoreit, Macketanz, Burchardt, Harris, Marheinecke, Rehm, Kayadelen, Attia, Elkahky, Yu, Pitler, Lertpradit, Mandl, Kirchner, Alcalde, Strnadov{\'a}, Banerjee, Manurung, Stella, Shimada, Kwak, Mendon{\c{c}}a, Lando, Nitisaroj, and Li}]{zeman-etal-2017-conll}
Daniel Zeman, Martin Popel, Milan Straka, Jan Haji{\v{c}}, Joakim Nivre, Filip Ginter, Juhani Luotolahti, Sampo Pyysalo, Slav Petrov, Martin Potthast, Francis Tyers, Elena Badmaeva, Memduh Gokirmak, Anna Nedoluzhko, Silvie Cinkov{\'a}, Jan Haji{\v{c}}~jr., Jaroslava Hlav{\'a}{\v{c}}ov{\'a}, V{\'a}clava Kettnerov{\'a}, Zde{\v{n}}ka Ure{\v{s}}ov{\'a}, Jenna Kanerva, Stina Ojala, Anna Missil{\"a}, Christopher~D. Manning, Sebastian Schuster, Siva Reddy, Dima Taji, Nizar Habash, Herman Leung, Marie-Catherine de~Marneffe, Manuela Sanguinetti, Maria Simi, Hiroshi Kanayama, Valeria de~Paiva, Kira Droganova, H{\'e}ctor Mart{\'\i}nez~Alonso, {\c{C}}a{\u{g}}r{\i} {\c{C}}{\"o}ltekin, Umut Sulubacak, Hans Uszkoreit, Vivien Macketanz, Aljoscha Burchardt, Kim Harris, Katrin Marheinecke, Georg Rehm, Tolga Kayadelen, Mohammed Attia, Ali Elkahky, Zhuoran Yu, Emily Pitler, Saran Lertpradit, Michael Mandl, Jesse Kirchner, Hector~Fernandez Alcalde, Jana Strnadov{\'a}, Esha Banerjee, Ruli Manurung, Antonio Stella, Atsuko Shimada,
  Sookyoung Kwak, Gustavo Mendon{\c{c}}a, Tatiana Lando, Rattima Nitisaroj, and Josie Li. 2017.
\newblock \href {https://doi.org/10.18653/v1/K17-3001} {{C}o{NLL} 2017 shared task: Multilingual parsing from raw text to {U}niversal {D}ependencies}.
\newblock In \emph{Proceedings of the {C}o{NLL} 2017 Shared Task: Multilingual Parsing from Raw Text to Universal Dependencies}, pages 1--19, Vancouver, Canada. Association for Computational Linguistics.

\bibitem[{Zhang et~al.(2019)Zhang, Li, Fu, and Zhang}]{zhang-etal-2019-syntax-enhanced}
Meishan Zhang, Zhenghua Li, Guohong Fu, and Min Zhang. 2019.
\newblock \href {https://doi.org/10.18653/v1/N19-1118} {Syntax-enhanced neural machine translation with syntax-aware word representations}.
\newblock In \emph{Proceedings of the 2019 Conference of the North {A}merican Chapter of the Association for Computational Linguistics: Human Language Technologies, Volume 1 (Long and Short Papers)}, pages 1151--1161, Minneapolis, Minnesota. Association for Computational Linguistics.

\bibitem[{Zhang et~al.(2018)Zhang, Qi, and Manning}]{zhang-etal-2018-graph}
Yuhao Zhang, Peng Qi, and Christopher~D. Manning. 2018.
\newblock \href {https://doi.org/10.18653/v1/D18-1244} {Graph convolution over pruned dependency trees improves relation extraction}.
\newblock In \emph{Proceedings of the 2018 Conference on Empirical Methods in Natural Language Processing}, pages 2205--2215, Brussels, Belgium. Association for Computational Linguistics.

\end{thebibliography}


\begin{thebibliography}{8}
\expandafter\ifx\csname natexlab\endcsname\relax\def\natexlab#1{#1}\fi

\bibitem[{Conneau et~al.(2019)Conneau, Khandelwal, Goyal, Chaudhary, Wenzek, Guzm{\'{a}}n, Grave, Ott, Zettlemoyer, and Stoyanov}]{xmlroberta}
Alexis Conneau and Kartikay Khandelwal and Naman Goyal and Vishrav Chaudhary and Guillaume Wenzek and Francisco Guzm{\'{a}}n and Edouard Grave and Myle Ott and Luke Zettlemoyer and Veselin Stoyanov. 2019.
\newblock \href {https://huggingface.co/xlm-roberta-base} {\emph{XLM-RoBERTa}}.
\newblock Hugging Face.

\bibitem[{Grave et~al.(2018)Grave, Bojanowski, Gupta, Joulin, and Mikolov}]{fasttext}
Grave, Edouard and Bojanowski, Piotr and Gupta, Prakhar and Joulin, Armand and Mikolov, Tomas. 2018.
\newblock \href {https://fasttext.cc/docs/en/crawl-vectors.html} {\emph{fastText}}.
\newblock Facebook.

\bibitem[{Kłeczek(2021)}]{polbert}
Kłeczek, Dariusz. 2021.
\newblock \href {https://huggingface.co/dkleczek/bert-base-polish-cased-v1} {\emph{Polbert}}.
\newblock Hugging Face.

\bibitem[{Lynn et~al.(2015)Lynn, Foster, McGuinness, Walsh, Phelan, and Scannell}]{IrishTreebank}
Lynn, Teresa and Foster, Jennifer and McGuinness, Sarah and Walsh, Abigail and Phelan, Jason and Scannell, Kevin. 2015.
\newblock \href {https://github.com/UniversalDependencies/UD_Irish-IDT} {\emph{{Irish Dependency Treebank (UD Irish-IDT)}}}.
\newblock Universal Dependencies Consortium.
\newblock PID \href{http://hdl.handle.net/11234/1-4611}{http://hdl.handle.net/11234/1-4611}.

\bibitem[{Mroczkowski et~al.(2021)Mroczkowski, Rybak, Wr{\'o}blewska, and Gawlik}]{herbert}
Mroczkowski, Robert and Rybak, Piotr and Wr{\'o}blewska, Alina and Gawlik, Ireneusz. 2021.
\newblock \href {https://huggingface.co/allegro/herbert-base-cased} {\emph{HerBERT}}.
\newblock Hugging Face.

\bibitem[{Przepiórkowski et~al.(2018)Przepiórkowski, Bańko, Górski, and Lewandowska-Tomaszczyk}]{NKJP1M}
Przepiórkowski, Adam and Bańko, Mirosław and Górski, Rafał L. and Lewandowska-Tomaszczyk, Barbara. 2018.
\newblock \href {http://clip.ipipan.waw.pl/NationalCorpusOfPolish?action=AttachFile&do=get&target=NKJP-PodkorpusMilionowy-1.2.tar.gz} {\emph{{National Corpus of Polish}}}.
\newblock Institute of Computer Science.

\bibitem[{Shen et~al.(2019)Shen, McDonald, Zeman, and Qi}]{ChineseTreebank}
Shen, Mo and McDonald, Ryan and Zeman, Daniel and Qi, Peng. 2019.
\newblock \href {https://github.com/UniversalDependencies/UD_Chinese-GSD} {\emph{{Chinese Dependency Treebank (UD Chinese-GSD)}}}.
\newblock Universal Dependencies Consortium.
\newblock PID \href{http://hdl.handle.net/11234/1-4611}{http://hdl.handle.net/11234/1-4611}.

\bibitem[{Wr{\'o}blewska(2018)}]{PDBUDdataset}
Wr{\'o}blewska, Alina. 2018.
\newblock \href {https://github.com/UniversalDependencies/UD_Polish-PDB} {\emph{{Polish Dependency Bank (UD Polish-PDB)}}}.
\newblock Universal Dependencies Consortium.
\newblock PID \href{http://hdl.handle.net/11234/1-5150}{http://hdl.handle.net/11234/1-5150}.

\end{thebibliography}

\section{Language Resource References}
\label{lr:ref}
\bibliographystylelanguageresource{lrec-coling2024-natbib}
\bibliographylanguageresource{languageresource}

\end{document}